\documentclass{WileyMSP-template}

\usepackage[pagewise]{lineno}
\usepackage{graphicx}%
\usepackage{multirow}%
\usepackage{amsmath,amssymb,amsfonts}%

\usepackage{amsthm}%
\usepackage{mathrsfs}%
\usepackage[title]{appendix}%
\usepackage{xcolor}%
\usepackage{textcomp}%
\usepackage{manyfoot}%
\usepackage{booktabs}%
\usepackage{algorithm}%
\usepackage{algorithmicx}%
\usepackage{algpseudocode}%
\usepackage{listings}%
\usepackage{makecell, booktabs}

\usepackage{amssymb}
\usepackage{multirow} 
\usepackage{adjustbox}
\usepackage{tabularx}
\newcolumntype{Y}{>{\centering\arraybackslash}X} 

\newcolumntype{L}{>{\raggedright\arraybackslash}X}

\begin{document}

\pagestyle{fancy}


\title{AI-Guided Human-In-the-Loop Inverse Design of High Performance Engineering Structures}

\maketitle


\author{Dat Quoc Ha*}
\author{Md Ferdous Alam}
\author{Markus J. Buehler}
\author{Faez Ahmed}
\author{Josephine V. Carstensen}


\dedication{}

\begin{affiliations}
D. Q. Ha, M. J. Buehler, J. V. Carstensen\\
Civil and Environmental Engineering \\
Massachusetts Institute of Technology \\
Cambridge, MA, USA \\
Email Address: datha@mit.edu, jvcar@mit.edu \\

\medskip

M. F. Alam, F. Ahmed\\
Mechanical Engineering \\
Massachusetts Institute of Technology \\
Cambridge, MA, USA \\

\end{affiliations}



\begin{abstract}
Inverse design tools such as Topology Optimization (TO) can achieve new levels of improvement for high-performance engineered structures. However, widespread use is hindered by high computational times and a black-box nature that inhibits user interaction. Human-in-the-loop TO approaches are emerging that integrate human intuition into the design generation process. However, these rely on the time-consuming bottleneck of iterative region selection for design modifications. To reduce the number of iterative trials, this contribution presents an AI co-pilot that uses machine learning to predict the user’s preferred regions. The prediction model is configured as an image segmentation task with a U-Net architecture. It is trained on synthetic datasets where human preferences either identify the longest topological member or the most complex structural connection. The model successfully predicts plausible regions for modification and presents them to the user as AI recommendations. The human preference model demonstrates generalization across diverse and non-standard TO problems and exhibits emergent behavior outside the single-region selection training data. Demonstration examples show that the new human-in-the-loop TO approach that integrates the AI co-pilot can improve manufacturability or improve the linear buckling load by 39\% while only increasing the total design time by 15 sec compared to conventional simplistic TO.
\end{abstract}



\section{Introduction}\label{sec:intro}

High-performance structures are essential for modern engineering. 
From aerospace components to offshore wind turbines and new light weight material architectures, they must balance material efficiency with structural integrity. 
Traditional design methods depend on manual iterations or predefined layouts, which often result in wasted materials or weakened performance. 
Fortunately, advances in computational power have revolutionized this process. 
Inverse design tools such as Topology Optimization (TO) now fully automates the search for optimal material distributions within a design space, subject to loads and constraints. 
By treating design as a mathematical problem, TO produces lightweight, efficient structures—free from the biases of human intuition. 
Successful examples of the application of TO can be found in numerous engineering disciplines. 
They include aerospace applications \cite{2016_zhu_aircraft,2011_tomlin_aircraft} where weight savings in aircrafts of up to 64\% have been reported \cite{2011_tomlin_aircraft}.
In mechanical engineering, TO has been used to achieve a 19\% increase in the efficiency of heat conducting structures \cite{2021_wu_thermal}, and to improve the mechanical performance of bell crank components in the F1 racing sidecar \cite{2020_pang_bellcrank}.
TO has been applied to improve several performance characteristics of low weight material architectures including energy absorption \cite{chen2018design,carstensen2022topology,2020_yang_lattice,2021_zhang_dome,dalklint2023computational,alkhatib2023isotropic,zeng2023inverse}.
It has also been used in electrical engineering to optimize antenna design, achieving up to 250\% bandwidth enhancement \cite{2003_kiziltas_dielectric}, and for numerous examples of performance improved photonic and phononic crystals \cite{2019_li_photonic}.
In civil engineering, TO has been has been shown to reduce the environmental impact of reinforced concrete construction by reducing the concrete material consumption by at least 25-30\% \cite{liu2020experimental,wethyavivorn2022topology,pressmair2023contribution}.

\begin{figure}
\centering
    \includegraphics[width = 0.75\textwidth]{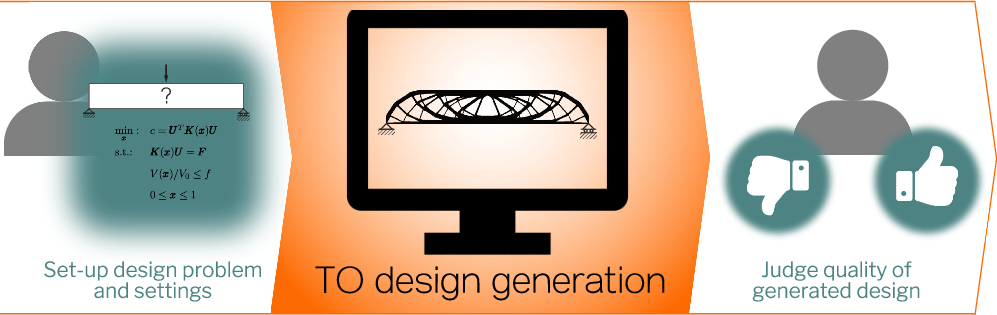} 
   \\
   (a) \\[15 pt]
    \begin{tabular}{ >{\centering\arraybackslash} m{0.125cm} >{\centering\arraybackslash} m{3.5in} >{\centering\arraybackslash} m{0.125cm} >{\centering\arraybackslash} m{3.5in} }
    (b.i) & \includegraphics[width=0.75\linewidth]{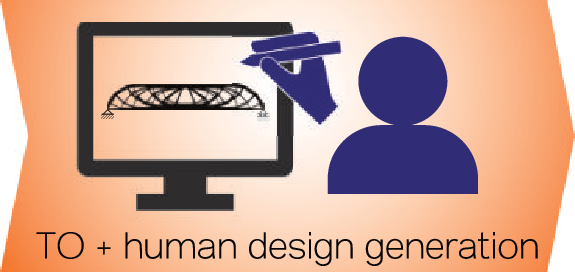} &
    (b.ii) & \includegraphics[width=0.75\linewidth]{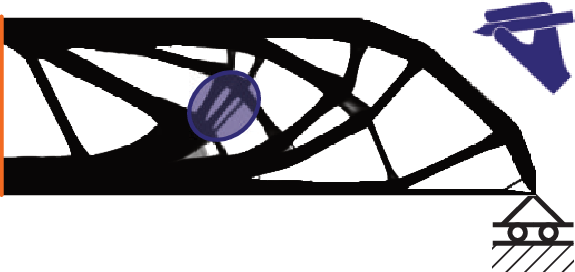} \\
    (b.iii) & \includegraphics[width=0.75\linewidth]{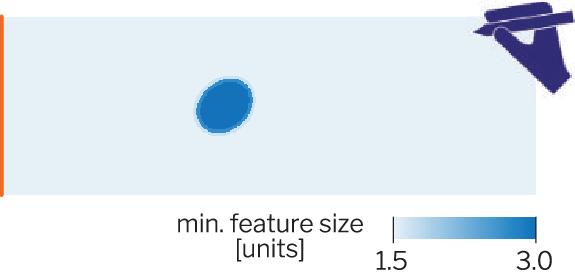} &
    (b.iv) & \includegraphics[width=0.75\linewidth]{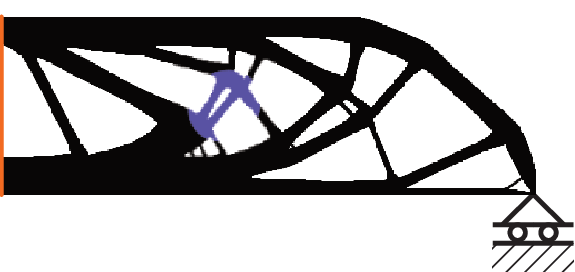} \\
\end{tabular} \\
 (b) \\[15 pt]
    \begin{tabular}{ >{\centering\arraybackslash} m{0.25cm} >{\centering\arraybackslash} m{7.5in} }
    (c.i) & \includegraphics[width=0.90\linewidth]{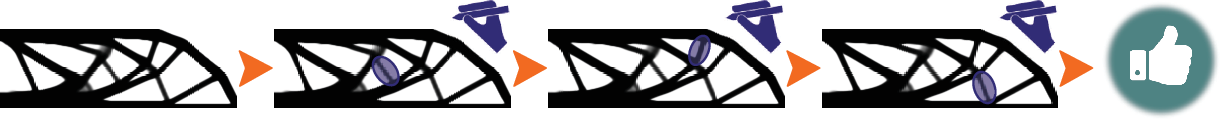} \\
    (c.ii) & \includegraphics[width=0.90\linewidth]{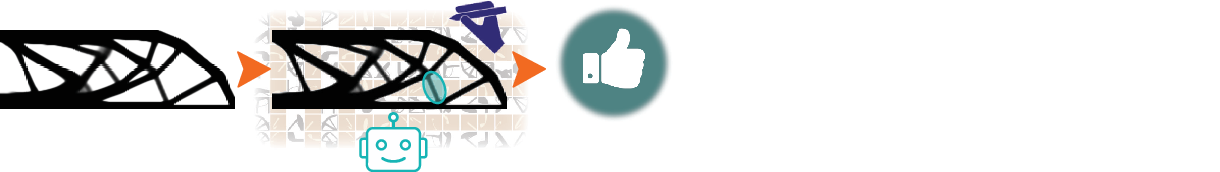} \\
        & \includegraphics[width=0.90\linewidth]{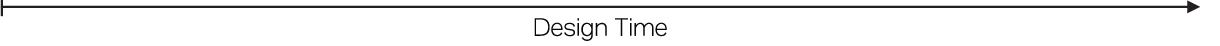} 
   \end{tabular}
   (c) \\[5 pt]
    \caption{(a) Classic TO where the human designer initiates the design process and a fully automated algorithm solves the design problem which the designer judges the final quality of. In contrast, (b) Human Informed Topology Optimization (HiTop) (b.i) allows TO to generate the design solution in \emph{collaboration} with the human user. In initially generates a quick automated design, and in (b.ii,b.iii), the human designer interactively updates the local minimum feature size requirements to fit their preferences, giving the new design in (b.iv). (c) Illustrates how the integration of an AI-co pilot will improve the design time for HiTop. In (c.i) the human uses multiple trials to identify the topological feature to modify, where as (c.ii) shows how a AI co-pilot recommends where modifications should be made based on previous interaction data. }
    \label{fig:hitop_MBB}
\end{figure}

Regardless of the engineering discipline, an inverse design task with TO is initiated by the design engineer. 
Figure \ref{fig:hitop_MBB}a left shows how the TO user starts the process by defining a physical space that material can occupy.
The user also needs to specify the anticipated loads and boundary conditions.
The design problem is formally formulated as a mathematical problem with an objective function and constraints.
The objective is usually a mechanical property that is measured by discretizing the physical space and solving a finite element problem. 
The constraint typically involves equilibrium with externally applied loads and limits on the available amount of material. 
The next step when using classic TO (Fig. \ref{fig:hitop_MBB}a center) is to solve the design problem with a fully automated algorithm.
When solving the TO problem, the algorithm seeks to find the distribution of material that maximizes or minimizes the objective while fulfilling the constraints.
This material distribution determines where both solid material and voids should be placed. 
Once a final material distribution is obtained, the design engineer evaluates the quality of the design and decides if redesign is needed (Fig. \ref{fig:hitop_MBB}a right), and if so, what modifications that needs to be made to the initial TO setup.

Despite the successful examples, classic TO has a key flaw for widespread implementation in engineering industries: it operates as a black box. 
As illustrated in Fig. \ref{fig:hitop_MBB}a, this means that the user is placed in a retroactive judgment stage, where they initialize the design and assess the output. 
However, most users want to combine the exploratory power of TO with their experience to improve more complex physical performance metrics or aesthetics that are not explicitly included in the design problem formulation \cite{loos2022towards,saadi2023generative,smith2024reducing}. Early works recognize that the black box nature of TO can be problematic and vaguely recommend that the designer's creativity should be included \cite{eschenauer2001topology}.
The notion has recently been reinforced by interviews with practicing designers from industrial and mechanical engineering as well as architectural design practices \cite{saadi2023generative}. The interviews reveal that designers iterate through the problem settings to create several design results to select from. The selected design is then subsequently refined to meet other requirements. One interviewee, e.g. reports: \emph{“I was also skeptical that these were not thick enough, so I made them thicker.”} 
However, several works have shown that these types of post-processing changes potentially compromise the performance of TO designs \cite{2016_lazarov,2019_jewett}.
Other survey results from structural engineering design practitioners and from aerospace, automotive, and steelmaking industries draw similar conclusions \cite{smith2024reducing,roy2008recent}. 
The studies reveal a general willingness to include automated design tools, including TO, more in the design practice, but that it is prevented by barriers related to \emph{“time,”} \emph{“learning curve,”} and \emph{“too black-box”}, motivating the need for design approaches that allow the user preferences to be actively leveraged as the design is generated. 

Some recent works addresses the black-box limitation using human-in-the-loop approaches that incorporate expert judgment directly into the TO workflow.
One example is Human-Informed Topology Optimization (HiTop) \cite{2023_hitop} that allows interactive adjustment of minimum feature sizes, enabling designers to improve outcomes for stress-sensitive or buckling-prone designs. 
The framework enables dynamic refinement of material distribution in critical areas, effectively combining algorithmic efficiency with engineering expertise. 
In HiTop, a simple TO problem formulation is used and interrupted after 50 iterations when the material distribution is giving an initial structural layout. 
On this initial material distribution, the user can interactively identify thin structural members which they wish to enlarge. A HiTop design example is shown in Fig. \ref{fig:hitop_MBB}b. In Fig. \ref{fig:hitop_MBB}b.ii, the user selects a central region with three thin structural members where they wish to double the minimum topological feature size from 1.5 units to 3.0 units. This modifications produce an updated feature size requirement map as shown in Fig. \ref{fig:hitop_MBB}b.iii. 
Re-optimization with the adjusted feature size requirements gives the final design in Fig. \ref{fig:hitop_MBB}b.iv.
It is seen that with the larger feature size requirement in the designated central region, the TO algorithm determines that maintaining three members here is structurally inefficient and therefore replaces them with two thicker elements.

Since the original publication of HiTop \cite{2023_hitop}, the interest in human-in-the-loop TO frameworks has grown as they have been shown only require short compute times and to improve complex performance metrics such as eigenvalues, linear buckling resistance, stress concentrations, energy absorption, manufacturing, and aesthetics. 
Recent works includes contributions that seeks to make TO more immersive to facilitate interactive selections in 3D.
Steps in this direction use immersive virtual reality (VR) environments either to improve visualization of design results \cite{2025_aragon} or to sculpt initial models \cite{2025_li_vrbeso} for a slightly different version of human-in-the-loop TO.
Other contributions include extensions to let the interactive user input to control both maximum and minimum length scales \cite{2023_gillian_hitop2}, or to enable control over structural topology and geometry parameters by blending implicit and explicit geometric descriptions \cite{2025_zhang_MMCSIMP}. 
Most recently, extensions using a two-pronged user input have been presented, where the user draws the changes they would like to see and where in the design they want them \cite{2023_li_besoscoring,2024_gillian_infill,schiffer2024integrating}. 

Although interactive TO approaches such as HiTop are much quicker than classic TO with complex constraints, their reliance on iterative human trial-and-error input can still be time-consuming.
As an example, the selection of regions to modify within the original HiTop framework is based on human intuition and often requires multiple iterative trials to identify the right region(s) for modifications (Fig. \ref{fig:hitop_MBB}c.i). 
This work stipulates that the desired human preferences to TO design are non-unique and that a AI co-pilot can speed up the iterative selection process by identifying the most likely region to change (Fig. \ref{fig:hitop_MBB}c.ii). 
Comprehensive reviews demonstrate significant progress in ML-TO integration \cite{2023_shin,2022_woldseth}, focusing mainly on iteration-free TO or on speeding up the evaluation of the design performance within each design iteration.  
Few studies have explored how ML can enhance human designer interactions. 
The current exceptions include approaches that predominantly employ sketch-based methods where sketches aid in guiding the TO iterations \cite{2025_zhu}, are interpreted into optimized structural configurations \cite{2021_garrelts}, or applied as constraints \cite{2024_zhang}. 
Recently, a unified approach to include geometric patterns and aesthetic constraints has also been proposed \cite{2026_tian_dpdnto}.
This contribution addresses a distinctly different challenge that has not been previously examined for high-performance inverse design in engineering; 
predicting structural areas that the user is likely to deem problematic. 
This work assumes that with sufficient data on user interactions, human preference patterns will emerge clearly for different classes of users that might emphasize different performance criteria, aesthetics, or manufacturing concerns.
Under this assumption, an AI augmentation will significantly accelerate the human design workflow by reducing the number of manual interventions while preserving the benefits of integrated expert judgment. 
The approach presented herein thus directly addresses the time-consuming bottleneck of manual, iterative region selection through an AI co-pilot that proactively predicts and recommends the user's preferred regions for modification.
It serves as a proof-of-concept for the technical framework with an hypothesized enhanced workflow efficiency.
Formal comparative studies of the new integrated framework against traditional methods are reserved for future work.

Within this contribution, we assume that when users select a region for modification on a TO design, their implicit goal is to isolate structural elements where they wish to increase the minimum feature size. 
This is a significant simplification of human design preferences, but here serve as the foundation for future, more involved AI-co pilots. 
The selection inherently divides the material distribution into two distinct classes. 
The first class comprises all elements contained within the selected region, while the second class includes all remaining material elements. 
Predicting a region to modify is thus essentially an image segmentation task.
In this work, we adapt similar segmentation principles to those originally developed for analyzing neuronal structures in electron microscopy data to the domain of topological feature identification \cite{2015_ronnebberger_unet}. 
The parallel between medical image analysis and topology segmentation proves particularly apt, as both require precise boundary detection and classification of structurally significant regions.
The new ML human preference model formulated herein learns from synthetic HiTop interaction data to predict user-preferred structural regions of concern. 
The ML model interprets and anticipates user decisions and it recommends regions for modifications through the AI co-pilot that is integrated in the new human-in-loop TO framework, HiTopAI. 
The designer's creative control is preserved by allowing the user to modify the AI co-pilots recommendations. 
The current contribution thus significantly advanced inverse design of high performance engineering components by addressing need for human-in-the-loop inclusions and short overall design workflow requirements.

\section{Results and Discussion}\label{sec:results_discussions}


\subsection{HiTop with AI Co-Pilot}\label{sec:result_mlhitop}



\begin{figure*}[h!]
    \centering
    \begin{tabular}{ccc}
    \includegraphics[scale = 1]{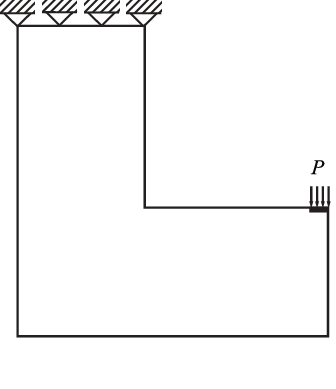} & 
    \includegraphics[scale = 1]{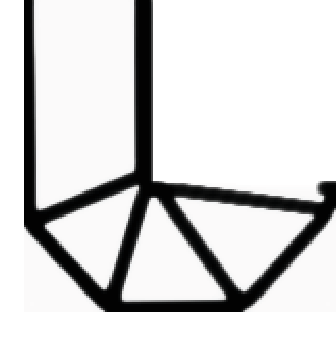} & 
 \includegraphics[scale = 1]{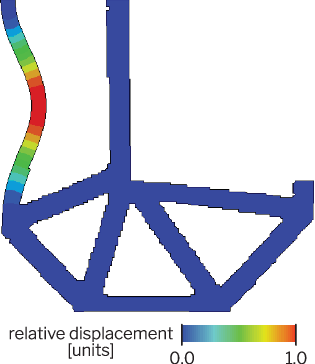} \\   
     (a) & (b) & (c) \\[5 pt] 
    \includegraphics[scale = 1]{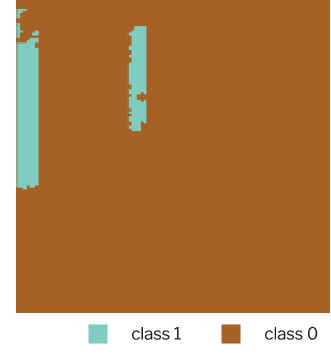} & 
    \includegraphics[scale = 1]{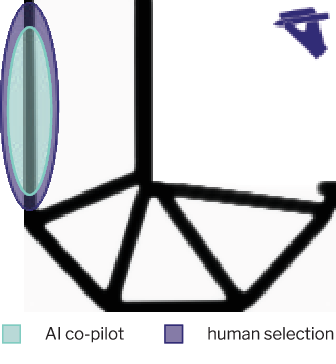} &
    \includegraphics[scale = 1]{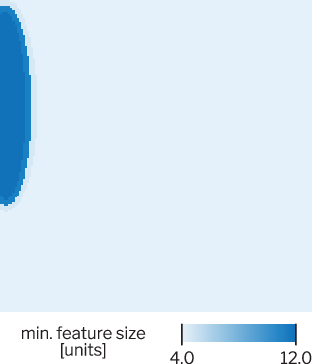} \\
        (d) & (e) & (f) \\[5 pt]  
    & \includegraphics[scale = 1]{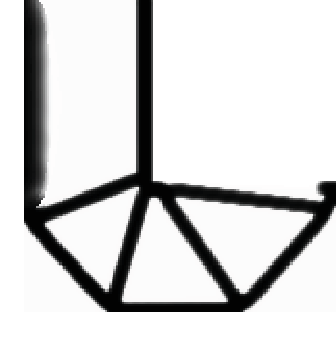} & 
    \includegraphics[scale = 1]{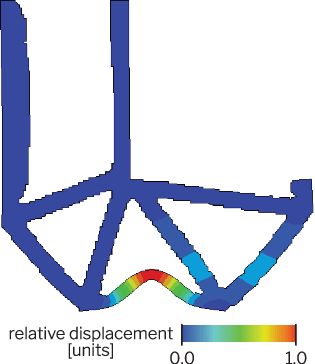} \\
    & (g) & (h)
    \end{tabular}
    \caption{Effect on buckling performance of using HiTopAI. The classic TO L-bracket design problem is shown in (a) which here is solved on a $150\times150$ unit$^2$ mesh, and (b) gives the initial topology after 50 iterations. The linear buckling performance for the initial design is $P_{cr}= 6.62\cdot10^{-3}P$ and the buckled shape is  shown in (c). 
    The post-processed ML prediction for the human preference region is given in (d), and recommended through the AI co-pilot as the cyan ellipse in (e). In (e), it is also illustrated how the user modifies the recommendation slightly to the purple ellipse. The user updates the minimum feature size within this region as shown in (f), which produces the final HiTopAI design in (g). The buckling performance of the final design is $P_{cr}= 9.20\cdot10^{-3}P$ and the buckled shape shown in (g). }
    \label{fig:demo_lbrac_problem} 
\end{figure*}

The new AI-assisted human-in-the-loop TO approach is initially demonstrated 
an example where we assume that the user will always seek to select and require modifications for the longest topological member. 
The selection of the longest topological member is chosen for ease of visualization; however as will be shown, the AI-co pilot can easily be extended to other features of interest such as the most complex node.
Figure \ref{fig:demo_lbrac_problem} illustrates the new workflow and how it can be used to quickly improve e.g. mechanical performance metrics.
We demonstrate this using an L-bracket design problem (Fig. \ref{fig:demo_lbrac_problem}a where the cutout is 60\% of the square domain size) with compliance minimization and a material restriction of 20\%.
Conventional TO results in an initial TO design with a clearly defined longest member (Fig. \ref{fig:demo_lbrac_problem}b). 
When the initial topology is fed as input to the trained ML model, it successfully segments the longest topological member seen as the largest class 1 (cyan) region on the model output in Fig. \ref{fig:demo_lbrac_problem}d. 
After post-processing of the model results where the largest connected cluster of class 1 pixels is identified, the smallest enclosing ellipse is constructed around it. The AI-co pilot recommends this ellipse as a region for user-modifications. 
The recommended ellipse is presented to the user instantaneously (0.1 sec) as the TO is interrupted, shown in cyan in Fig. \ref{fig:demo_lbrac_problem}e. 
For this example, the initial recommendation (cyan ellipse in Fig. \ref{fig:demo_lbrac_problem}e) requires slight user adjustment to fully encapsulate the target feature.
The user spends 15 sec on making these adjustments and finalizes the selection (purple ellipse).
The user specifies a new minimum feature size within this region that is increased from 4 to 12 units (Fig. \ref{fig:demo_lbrac_problem}f). 
The HiTopAI framework subsequently proceeds by incorporating the selections. 
The TO is rerun with the updated, larger feature size requirement, yielding a final design where the longest member exhibits increased thickness (Fig. \ref{fig:demo_lbrac_problem}g).
Material redistribution in other areas of the topology also occurs to accommodate the modified design requirements. 
The entire HiTopAI design process is completed in just under 6 minutes (344 sec).
To demonstrate the effect of human intervention, the linear buckling performance of both the initial and final TO designs is analyzed using Abaqus 2025.
The analyses reveal that the longest member (which is in compression) buckles first in the initial TO design (Fig. \ref{fig:demo_lbrac_problem}c).
The new HiTopAI design shows a different buckling mode (Fig. \ref{fig:demo_lbrac_problem}h) with a $39\%$ higher buckling load compared to the initial design.
The total design time to achieve this improvement is 344 sec compared to 329 sec for full convergence with conventional TO and thus only a 4\%

\begin{figure*}[h!]
    \centering
    \begin{tabular}{ccc}
    \includegraphics[scale=1]{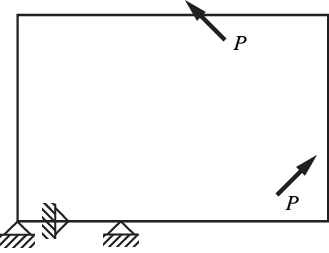} & 
    \includegraphics[scale=1]{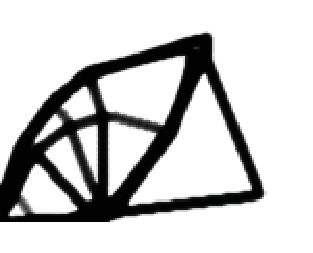} & \includegraphics[scale=1]{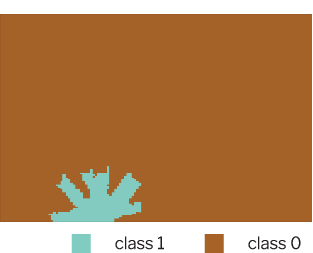}
    \\
    (a) & (b) & (c) \\ 
    \includegraphics[scale=1]{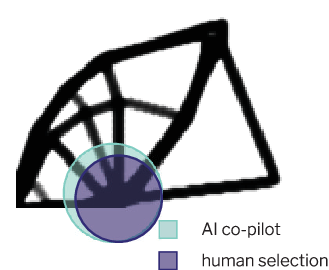} & 
    \includegraphics[scale=1]{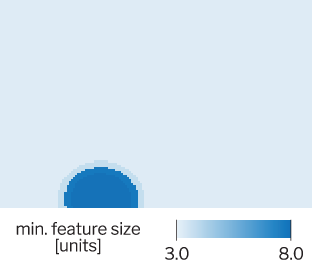} & \includegraphics[scale=1]{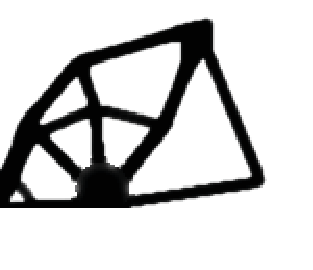} \\
    (d) & (e) & (f)
    \end{tabular}
    \caption{Example using HiTopAI to modify the most complex node. (a) gives the design problem that is solved on a $150\times 100$ unit$^2$ mesh. (b) shows the initial TO result after 50 iterations. The ML model's output is shown in (c), which displays the post-processed segmentation mask. The co-pilot's suggested intervention is represented by the cyan ellipse in (d), which the human user then fine-tunes into the purple ellipse. (e) illustrates the resulting minimum feature size field updated by the human intervention, leading to the final optimized design with reduced node complexity shown in (f).}
    \label{fig:demo_mostcomplexnode} 
\end{figure*}

To demonstrate the generalizability of the HiTopAI framework beyond human selection of the longest topological member, we introduce a second preference selection criterion in the AI-co pilot.
For this example, we focus on human preferences that limit node complexity as this is relevant for several manufacturing processes.
Herein, we define and quantify the complexity of a node by how many topological members it connects.
A complex design case (taken from the TopOpt game \cite{2016_nobeljorgensen_togame}) is shown in Fig. \ref{fig:demo_mostcomplexnode}a and Fig. \ref{fig:demo_mostcomplexnode}b gives the initial TO design. 
The most complex node is located at the right boundary condition and connects six topological members. 
Some of the connected members must be joined with small angular spacing (smallest at \textcolor{black}{20.4}\textcolor{black}{$^\circ$), which, depending on the planned fabrication process, can pose additional challenges for manufacturability. 
The initial topology is used as input for a ML model trained on identifying the most complex node, and the output segmentation in Fig. \ref{fig:demo_mostcomplexnode}c shows that this successfully achieved. 
As before, a post-processing step extracts the smallest bounding ellipse around the largest continuous cluster of predicted pixels, presenting the user with the cyan AI co-pilot recommended region in Fig. \ref{fig:demo_mostcomplexnode}d. 
Again, the user slightly modifies the region to the purple ellipse where the minimum feature size is increased (Fig. \ref{fig:demo_mostcomplexnode}e).
The final design is shown in Fig. \ref{fig:demo_mostcomplexnode}f where the human-in-the-loop design approach is observed to directly reduce the node complexity. 
By increasing the minimum feature size in the immediate vicinity of the complex node, the structural members that join at the node are thickened. This forces the TO algorithm to simplify the connection. 
The number of members that connect at the node is reduced to five and the angular spacing is increased such that the minimum spacing now is} 40.9$^\circ$.
The new design appears less complex and thus has improved manufacturabiltiy for most conventional fabrication processes.

\subsection{Performance of the Human Preference Prediction Model}
The time saving potential of the integrated AI co-pilot hinges on the performance of the human preference prediction model. 
Since the human user is still in-the-loop and able to make changes to the recommended region for design intervention, it should be emphasized that good model predictions are necessary, but \emph{perfect} model predictions are not. 
We have observed similar trends for both herein considered human preferences of the longest topological member and most complex node, respectively. 
Consequently, the discussion herein will initially focus on modeling the longest topological member, before extending to node complexity. 
As will be detailed in Section \ref{sec:data_creation}, the dataset that is used to model the selection of the longest topological member consists of 109,722 data pairs of square TO designs and synthetically generated human selections. The model is trained on 70\% of the dataset (76,806 data pairs). 
The training takes 3.8 hours, where 10\% of the dataset is used for validation loss as part of the termination criterion.
The remaining 20\% is used for evaluating the model performance after training.
All human selections in the dataset only identify a \emph{single} region for making human design interventions.
The model's input is a TO design without selection, and it's ability to segment either the longest topological member or the most complex node, serves herein as a critical test of its output quality.
After observing that the trained model has difficulty in identifying features that are on the boundary of the design domain, we pre-process all model input with a 10-pixel void border padding. This ensures all boundary features are surrounded by void and enables the model to successfully capture these as well. 

\subsubsection{Unseen data from the test dataset}\label{sec:result_test}
\begin{figure}
    \flushright
     \includegraphics[scale = 1.1]{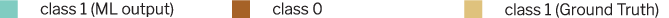} \\[5 pt] 
    \centering
    \begin{tabular}{ >{\centering\arraybackslash} m{0.25cm} >{\centering\arraybackslash} m{7.75in} }
(a.i) &
\includegraphics[scale = 0.85]{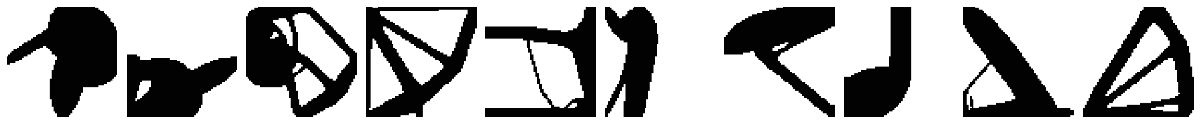} \\
(a.ii) & \includegraphics[scale = 0.85]{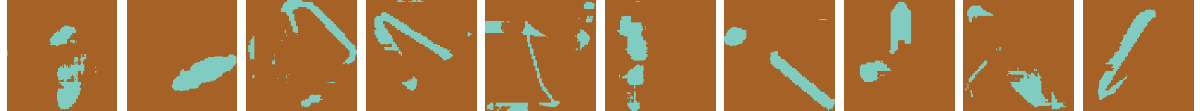} \\
(a.ii) & \includegraphics[scale = 0.85]{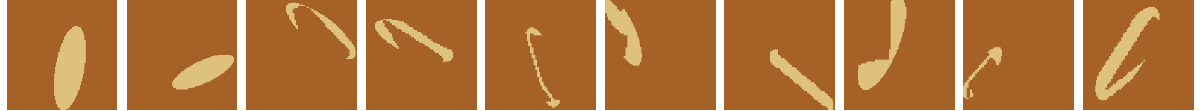} \\
\end{tabular}
\centering
(a) \\[15 pt]
    \flushright
     \includegraphics[scale = 1.1]{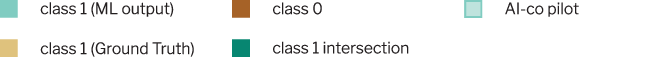} \\
    \centering
    \begin{tabular}{m{1.1 in} 
    >{\centering\arraybackslash} m{0.5in} 
    >{\centering\arraybackslash} m{0.5in} 
    >{\centering\arraybackslash} m{0.5in}
    >{\centering\arraybackslash} m{0.01 in} 
    >{\centering\arraybackslash} m{0.5in} 
    >{\centering\arraybackslash} m{0.5in} 
    >{\centering\arraybackslash} m{0.5in}
    >{\centering\arraybackslash} m{0.01 in} 
    >{\centering\arraybackslash} m{0.5in} 
    >{\centering\arraybackslash} m{0.5in} 
    >{\centering\arraybackslash} m{0.5in}}\\
    \multirow[c]{1}{1.1in}{TO Design} & 
    \includegraphics[width = 0.5in]{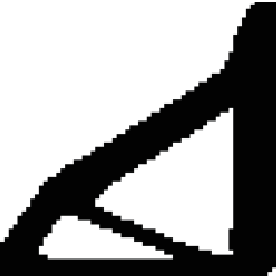} & 
    \includegraphics[width = 0.5in]{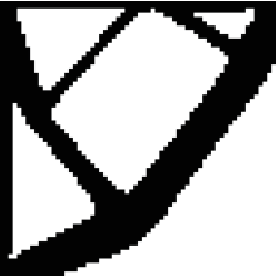} & 
    \includegraphics[width = 0.5in]{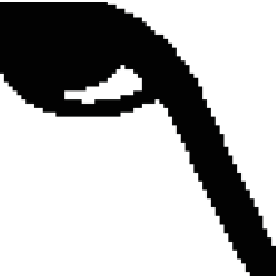} &
    &
    \includegraphics[width = 0.5in]{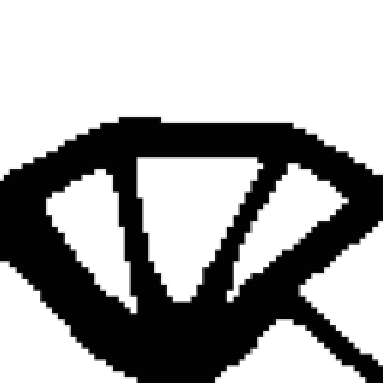} & 
    \includegraphics[width = 0.5in]{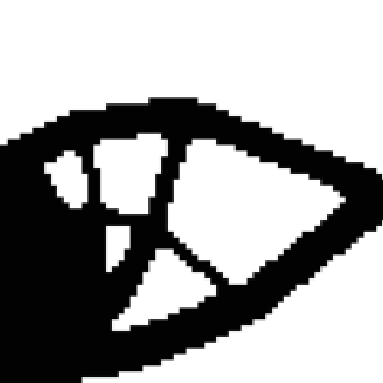} & 
    \includegraphics[width = 0.5in]{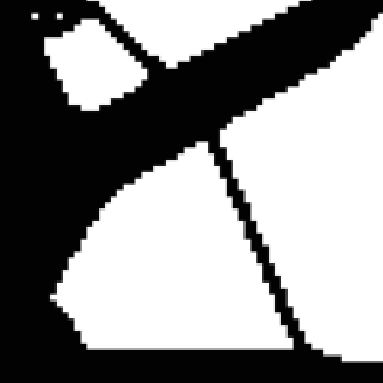} &
    &
    \includegraphics[width = 0.5in]{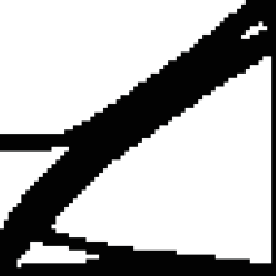} & 
    \includegraphics[width = 0.5in]{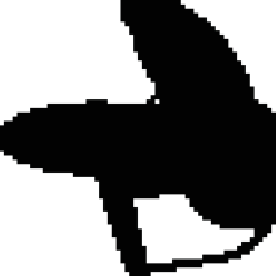} & 
    \includegraphics[width = 0.5in]{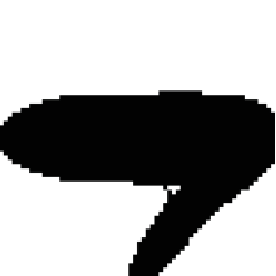} \\
    ML Output & 
    \includegraphics[width = 0.5in]{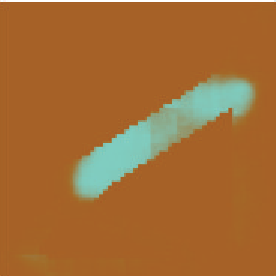} & 
    \includegraphics[width = 0.5in]{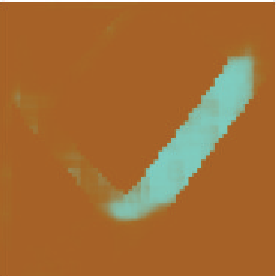} & 
    \includegraphics[width = 0.5in]{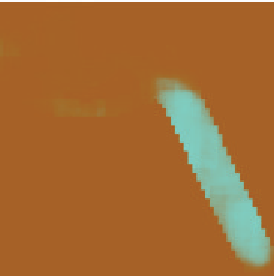} &
    &
    \includegraphics[width = 0.5in]{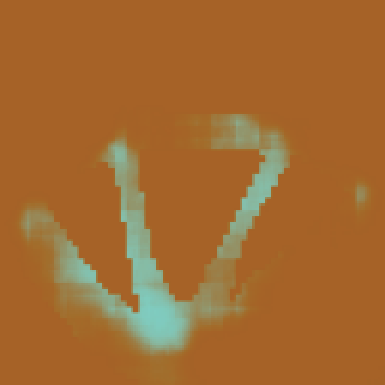} & 
    \includegraphics[width = 0.5in]{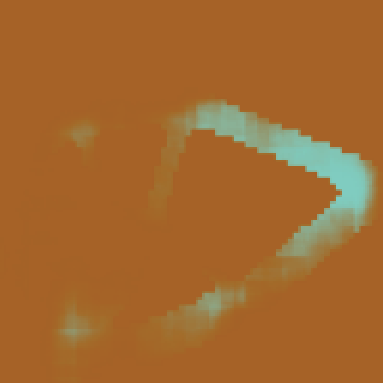} & 
    \includegraphics[width = 0.5in]{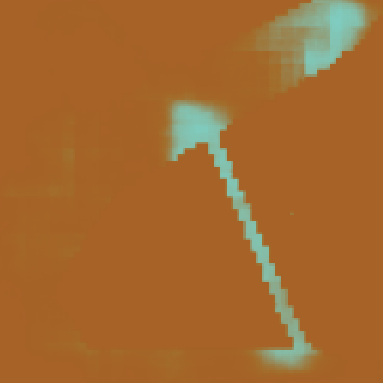} &
    &
    \includegraphics[width = 0.5in]{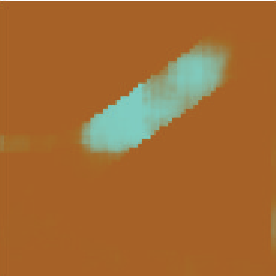} & 
    \includegraphics[width = 0.5in]{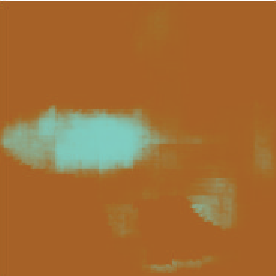} & 
    \includegraphics[width = 0.5in]{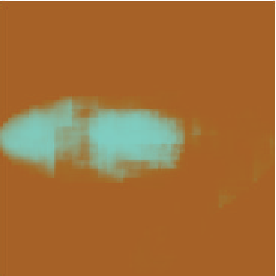} \\
    ML Output (Post-Processed) & 
    \includegraphics[width = 0.5in]{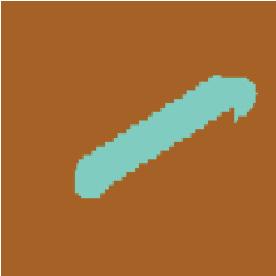} & 
    \includegraphics[width = 0.5in]{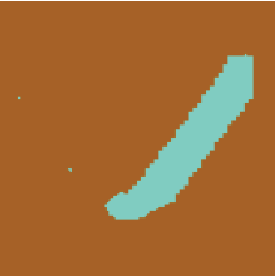} & 
    \includegraphics[width = 0.5in]{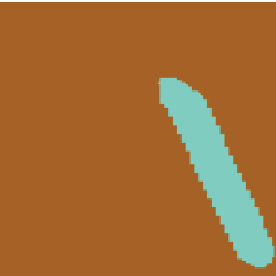} &
    &
    \includegraphics[width = 0.5in]{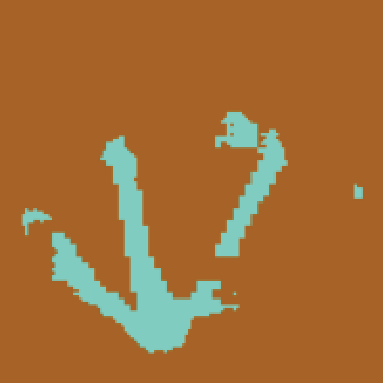} & 
    \includegraphics[width = 0.5in]{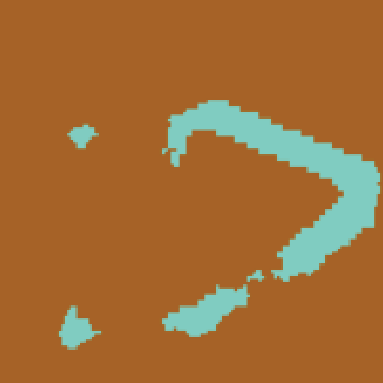} & 
    \includegraphics[width = 0.5in]{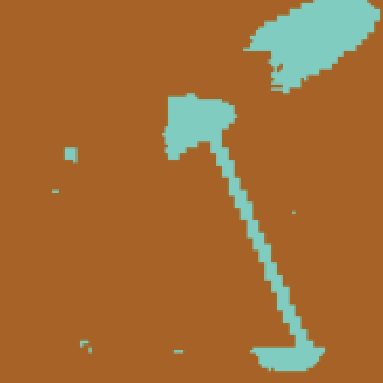} &
    &
    \includegraphics[width = 0.5in]{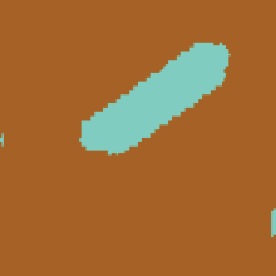} & 
    \includegraphics[width = 0.5in]{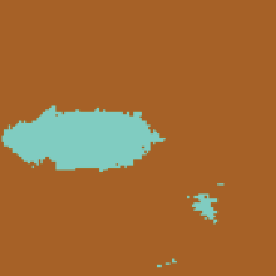} & 
    \includegraphics[width = 0.5in]{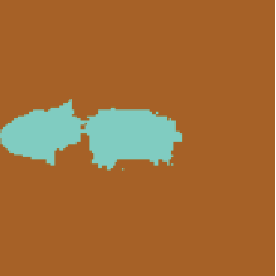} \\
    AI co-pilot's Recommendation & 
    \includegraphics[width = 0.5in]{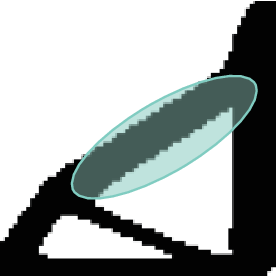} & 
    \includegraphics[width = 0.5in]{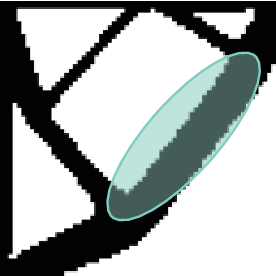} & 
    \includegraphics[width = 0.5in]{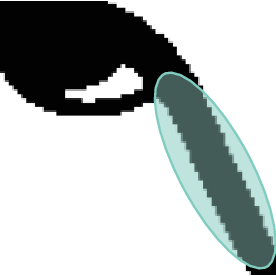} &
    &
    \includegraphics[width = 0.5in]{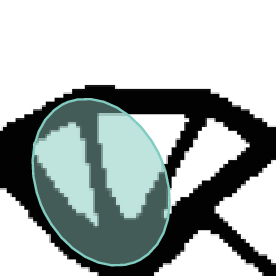} & 
    \includegraphics[width = 0.5in]{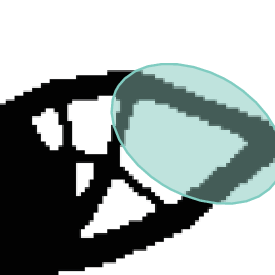} & 
    \includegraphics[width = 0.5in]{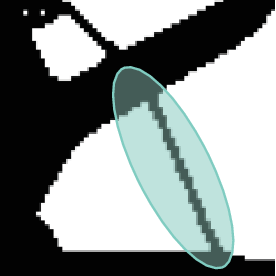} &
    &
    \includegraphics[width = 0.5in]{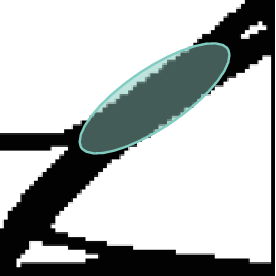} & 
    \includegraphics[width = 0.5in]{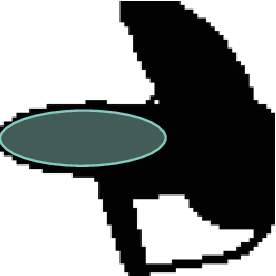} & 
    \includegraphics[width = 0.5in]{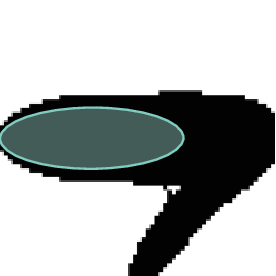} \\[5 pt]
    Comparison of ML and Ground Truth & 
    \includegraphics[width = 0.5in]{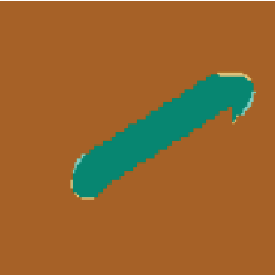} & 
    \includegraphics[width = 0.5in]{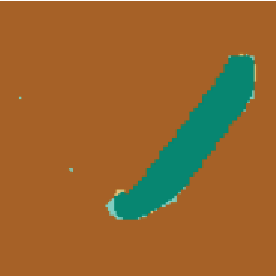} & 
    \includegraphics[width = 0.5in]{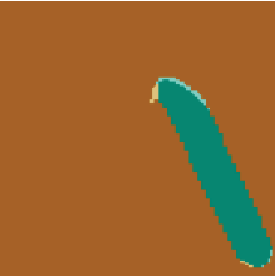} &
    &
    \includegraphics[width = 0.5in]{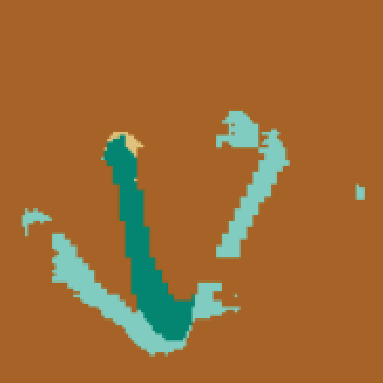} & 
    \includegraphics[width = 0.5in]{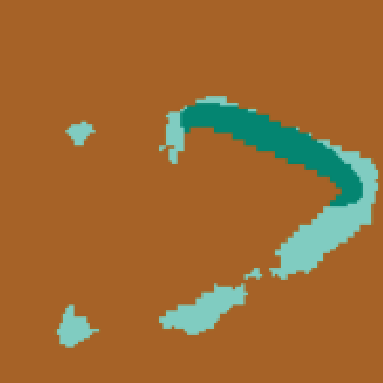} & 
    \includegraphics[width = 0.5in]{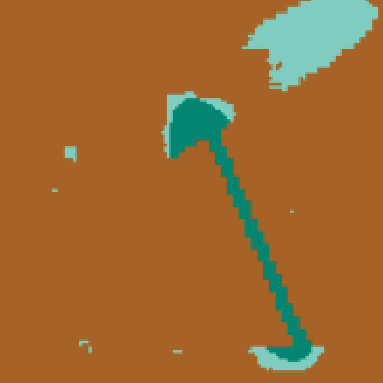} &
    &
    \includegraphics[width = 0.5in]{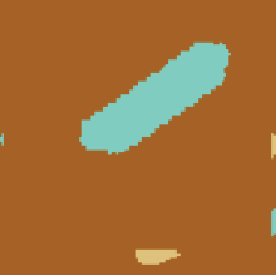} & 
    \includegraphics[width = 0.5in]{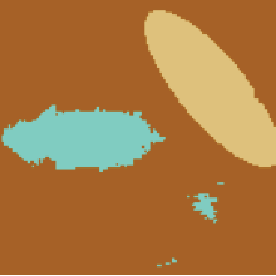} & 
    \includegraphics[width = 0.5in]{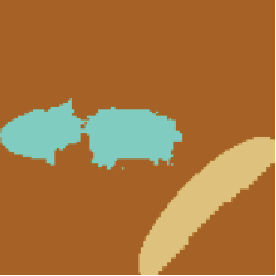} \\[5 pt]
    IOU & 
    0.97 & 0.97 & 0.97 &
    & 
    0.38 & 0.38 & 0.38 &
    &
    0.00 & 0.00 & 0.00 \\[5 pt]   
     & \multicolumn{3}{c}{(b.i)} &
     & \multicolumn{3}{c}{(b.ii)} &
     & \multicolumn{3}{c}{(b.iii)} \\ 
   \end{tabular}
   \centering 
   (b) 
    \caption{Performance of the human preferences ML model on unseen TO designs from the test dataset. (a) gives examples of the model's performance on 10 randomly designs. The input topologies are given in (a.i), (a.ii) shows the ML model's output (post-processed), and (a.iii) provides the corresponding ground truths. 
    In (b), examples of IOU metric trends observed for the trained ML model of human preference predictions is shown. 
    The first case (b.i) sees the most desirable IOU performance with IOU values of 0.97 which occurs for topologies with a clear longest member. The second case (b.ii), with median IOU values of 0.38, shows that the model's thresholded predictions do not fully match, but tends to overestimate compared to the ground truth. Finally, the third case (b.iii), with IOU values of close to 0, indicates that the synthetically generated dataset is imperfect. }
    \label{fig:mlhitop_testdataset_results}
\end{figure}

After training is complete, we first evaluate the model using unseen data from our test dataset (20\% of the total dataset). 
Figure \ref{fig:mlhitop_testdataset_results}a shows the model's prediction of the longest topological member for 10 random test samples. 
The model takes an input topology (Fig. \ref{fig:mlhitop_testdataset_results}a.i) and predicts pixel-wise probabilities for class 1 (human selection regions) versus class 0 (non-human selection regions). 
Applying a 90th percentile threshold converts these probability maps into binary segmentation masks as shown in Fig. \ref{fig:mlhitop_testdataset_results}a.ii.
Comparison with ground truth data for the same topologies (Fig. \ref{fig:mlhitop_testdataset_results}a.iii) reveals the model's strong capability to identify the longest topological member, though with some limitations. 
Imperfections appear as stray pixels in predictions (visible in samples 1,4,8, and 10), necessitating post-processing for practical integration as a co-pilot in HiTop. 
These observations also provide insights into dataset quality, further discussed in Section \ref{sec:data_creation}.

The intersection over union (IOU) measurement is used to quantitatively assess the performance of the trained human preference model on unseen data from the test dataset.
The IOU assesses the degree of agreement between the ML model's prediction and the ground truth. 
The best possible IOU score is $1$, where all the pixels with class 1 from the model prediction match with the ground truth.
For the model trained to detect the longest topological member, the average IOU score over the \textcolor{blue}{21,944} unseen data pairs is 0.58.
When switching the human preference to the most complex node, the IOU over the same amount of unseen data pairs is 0.45.
While far from a perfect $1$, this score reveals that the model starts to identify the features of interest for a given topology.
Inspecting the IOU score for individual test topologies, some patterns can be identified that help to explain the modest average IOU score and how its effect on the performance of the AI co-pilot is less pronounced than the numerical value indicates.
Figure \ref{fig:mlhitop_testdataset_results}b shows some of the different behaviors observed of the model trained to detect the longest topological member on unseen data from the test dataset.
The first case (Fig. \ref{fig:mlhitop_testdataset_results}b.i) achieves the most desirable performance with an average IOU of 0.97. 
This occurs when the input topologies features a clearly defined longest member. As seen in the three shown examples, the intersection of the ML model's prediction and the ground truth is near perfect.
Since the AI co-pilot's recommendation is based on the largest cluster of pixels in the ML model, this also aligns near perfectly with the ground truth.  
Across the test dataset, only 2.9\% of the model outputs have IOU values above 0.80.
The second case (Fig. \ref{fig:mlhitop_testdataset_results}b.ii), with IOU values around 0.38 as the median, demonstrates a partial mismatch between the model's thresholded predictions and the ground truth. 
This is the most frequently observed model performance, with 38\% of the test dataset achieving IOU scores in the range 0.3-0.5.
The discrepancy tends to appear in topologies that have multiple candidate members of similar length.
The post-processed ML model here predicts multiple long members, while the current dataset annotation only labels a single longest member per topology. 
For all three examples in Fig. \ref{fig:mlhitop_testdataset_results}b.ii, the ground truth is contained within the ML's models prediction of the largest cluster of pixels.
The AI co-pilot's recommendation thus contains the ground truth and can quickly be adjusted by the user to make up for the low IOU scores. 
Such adjustments would likely be made by the user for the examples in columns 1 and 2 of Fig. \ref{fig:mlhitop_testdataset_results}b.ii.
The AI co-pilot's recommendation is closer to the ground truth than the ML model's prediction in column 3 and it's quality is thus unaffected by the low IOU score. 
The third case (Fig. \ref{fig:mlhitop_testdataset_results}b.iii) has IOU values near 0 and highlights some limitations of the synthetic dataset generation used herein. 
The figure gives some of the worst examples, but 16.0\% of the test dataset achieves IOU values at or below 0.20.
As will be detailed in Section \ref{sec:skeletonization}, an automatic skeletonization process is used to identify the longest topological member. The method used sometimes misidentifies the longest member as seems to be the case for the ground truth in the first column of Fig. \ref{fig:mlhitop_testdataset_results}b.iii.
Identification of topological members trough skeletonization is especially difficult for topologies with thick and sparse structural elements as is the case in columns 2 and 3 where we also observe that the model's selected topological features completely diverges from the ground truth. 
Since the ML's model results only serves the foundation for a \emph{recommendation} for where to make changes, the human user still has the ability to reject or significantly modify this region before proceeding.

\subsubsection{Generalizability of the model}\label{sec:result_general}

\begin{figure*}[h!]
    \flushright
     \includegraphics[scale = 1.2]{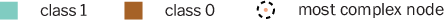} \\[5 pt] 
    \centering
    \begin{tabular}{cccc}
    \includegraphics[height = 0.15\linewidth]{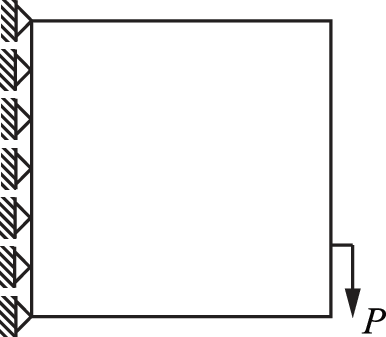} & 
    \includegraphics[height = 0.15\linewidth]{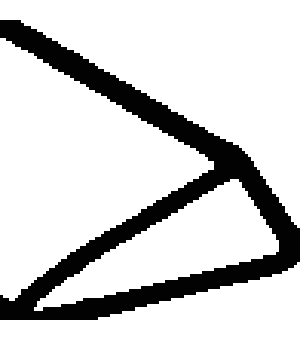} 
    & \includegraphics[height = 0.15\linewidth]{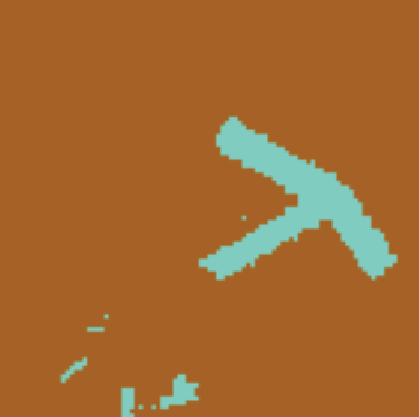} & \includegraphics[height = 0.15\linewidth]{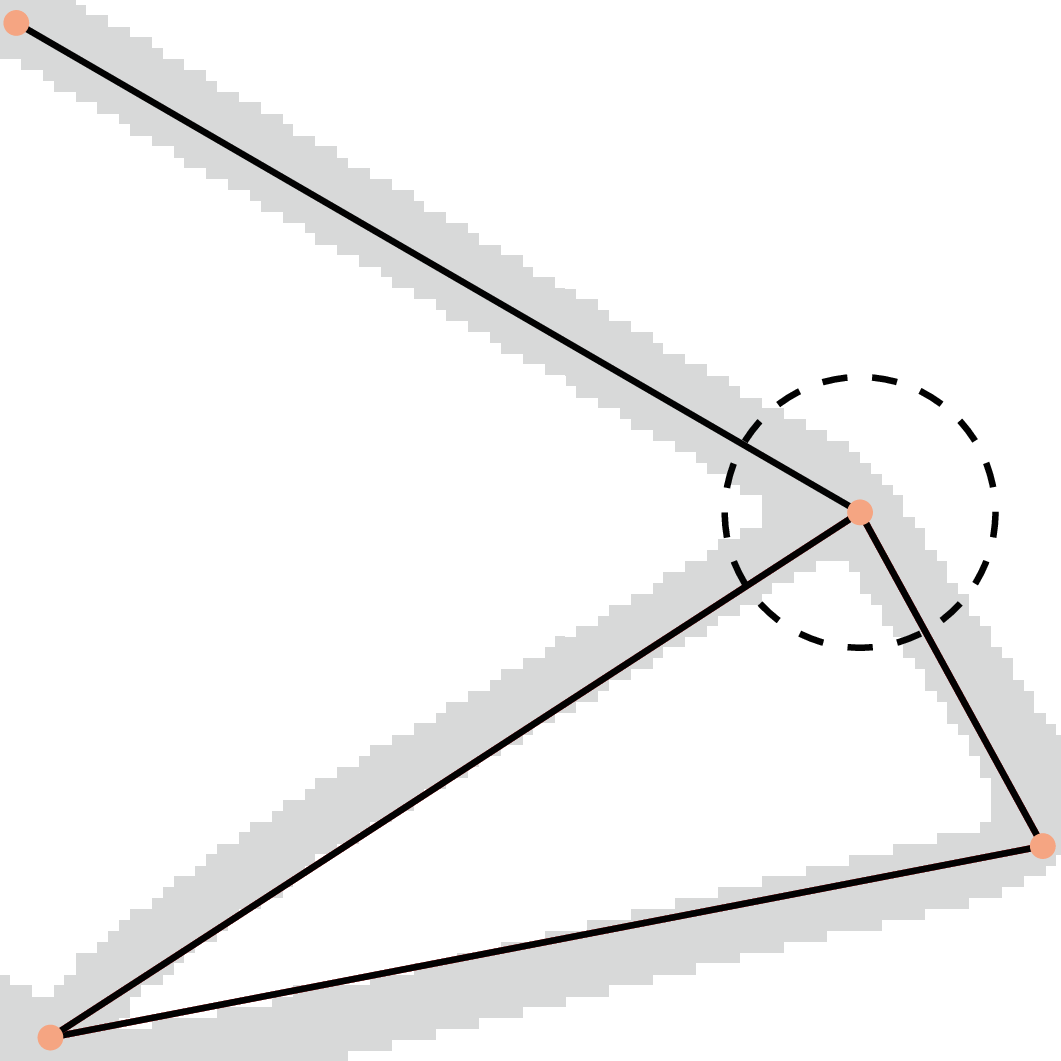} \\
    (a) & (b) & (c) & (d)\\[5 pt] 
    \includegraphics[height = 0.15\linewidth]{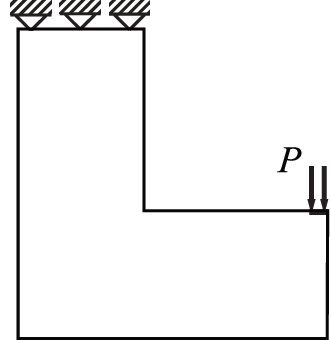} & 
    \includegraphics[height = 0.15\linewidth]{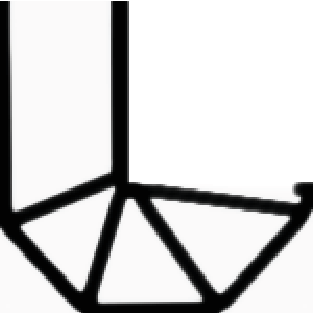} & \includegraphics[height = 0.15\linewidth]{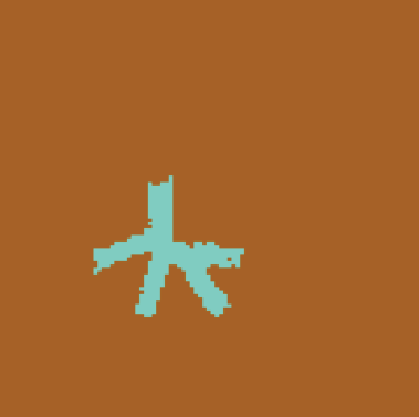} & \includegraphics[height = 0.15\linewidth]{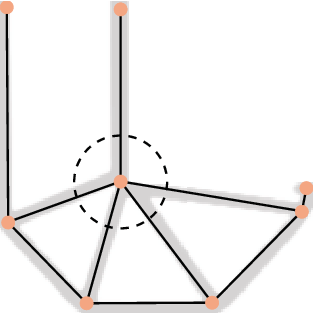} \\
    (e) & (f) & (g) & (h)\\[5 pt]  
    \includegraphics[height = 0.15\linewidth]{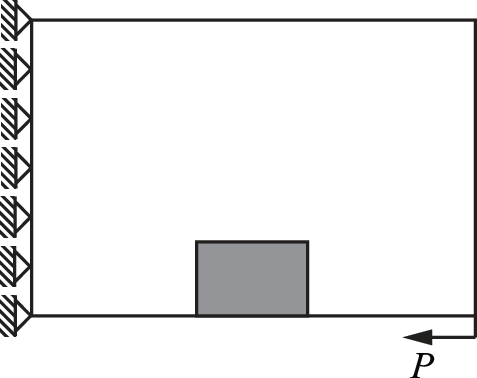} & 
    \includegraphics[height = 0.15\linewidth]{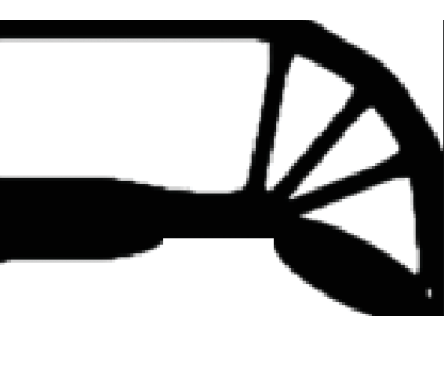} & \includegraphics[height = 0.15\linewidth]{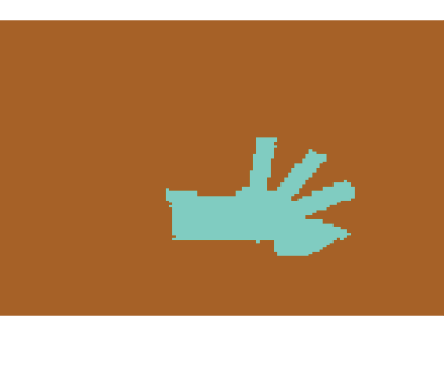} & \includegraphics[height = 0.15\linewidth]{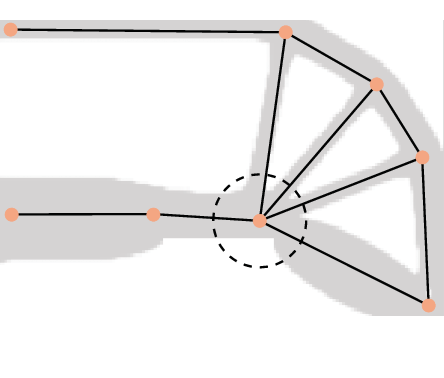} \\
    (i) & (j) & (k) & (l)\\
    \end{tabular}
    \caption{Examples of the performance of the trained ML model's predictions on TO designs outside the training dataset. Here the model is trained on segmenting the most complex node. Three TO problems are show in (a, e, i), and the corresponding initial TO designs after 50 iterations are given in (b, f, j).
    The post-processed ML modeling results for human preference for interventions are show as segmentation masks in (c, g, k). For comparison, skeletonized representations of the initial TO designs are given in (d, h, l), with the most complex node highlighted by a dashed line circle.}
    \label{fig:mlhitop_general_mostcomplexnode} 
\end{figure*}

Figure \ref{fig:mlhitop_general_mostcomplexnode} demonstrates the model's segmentation performance on TO designs that are not included in the test dataset.
For these examples, the human preference model is trained on data that identifies the most complex structural connection.
The figure shows the design domain for three demonstration problems in Fig. \ref{fig:mlhitop_general_mostcomplexnode}a,e,i and the resulting TO designs without human intervention in Fig. \ref{fig:mlhitop_general_mostcomplexnode}b,f,j. The TO designs are used as input for the trained ML that outputs the segmentation predictions in Fig. \ref{fig:mlhitop_general_mostcomplexnode}c,g,k. 
To ease visual confirmation of the model's results, skeletonization is overlaid on the TO designs in Fig. \ref{fig:mlhitop_general_mostcomplexnode}d,h,l where the most complex structural connections are indicated with dashed circles.

The first design case (Fig. \ref{fig:mlhitop_general_mostcomplexnode}a-d) examine the classic benchmark TO cantilever using 1:1 aspect ratio domains as in the test dataset. 
The load is applied at one-quarter of the domain height from the bottom right corner (Fig. \ref{fig:mlhitop_general_mostcomplexnode}a), resulting in a TO design where the most complex node joins three structural elements (Fig. \ref{fig:mlhitop_general_mostcomplexnode}b).
The trained ML model successfully segments the this node (Fig. \ref{fig:mlhitop_general_mostcomplexnode}c).
Note that the trained ML model has no access to the skeletonization data in Fig \ref{fig:mlhitop_general_mostcomplexnode}d during inference. 

The model is also tested on the L-bracket problem and non-standard TO design problems from the TopOpt game \cite{2016_nobeljorgensen_togame}, that features unconventional TO design challenges. 
Unlike the training data, some of these have design domains that are not square, but have aspect ratios of upto 1.5:1.
For the design challenges in Fig. \ref{fig:mlhitop_general_mostcomplexnode}e,i, the corresponding TO design results (Fig. \ref{fig:mlhitop_general_mostcomplexnode}f,j) have clearly defined complex connections that in both cases join five structural members (Fig. \ref{fig:mlhitop_general_mostcomplexnode}h,l).
Again, the trained ML human preference prediction model outputs show correctly identified and segmentation of the most complex connections (Fig. \ref{fig:mlhitop_general_mostcomplexnode}g,k). 

\begin{figure*}[h!]
    \flushright
     \includegraphics[scale = 1.2]{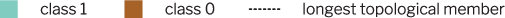} \\[5 pt] 
    \centering
    \begin{tabular}{cccc}
    \includegraphics[scale = 0.85]{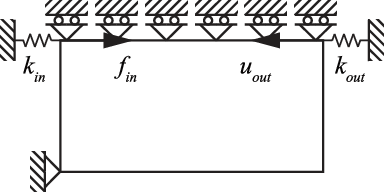} & 
    \includegraphics[scale = 0.85]{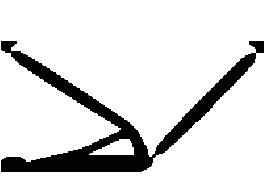} & \includegraphics[scale = 0.85]{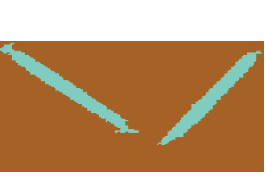} & \includegraphics[scale = 0.85]{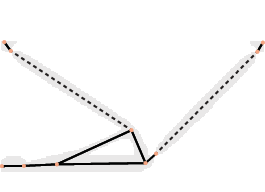} \\
    (a) & (b) & (c) & (d)\\[5 pt] 
    \includegraphics[scale = 0.85]{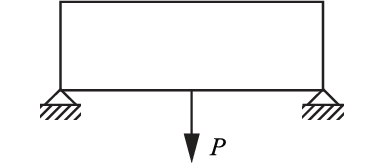} & 
    \includegraphics[scale = 0.85]{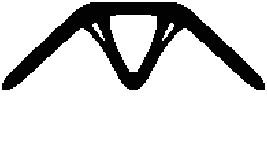} & \includegraphics[scale = 0.85]{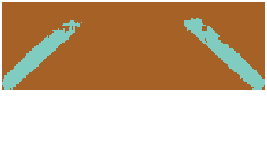} & \includegraphics[scale = 0.85]{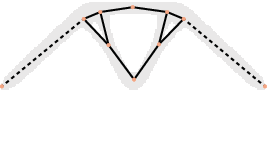} \\
    (e) & (f) & (g) & (h)
    \end{tabular}
    \caption{The ML model is trained on data with a \emph{single} human preference selection, but shows ability to segment multiple members in TO designs where two structural elements have almost the same maximum length. Two TO problems are given in (a, e) and their initial topologies are shown in (b, f). The resulting segmentations (c, g) demonstrate the model's identification of dual candidate members. For comparison, (d, h) illustrates the skeletonization of the TO designs, highlighting with dashed lines both the longest member and members within 10 \% of this length.}
    \label{fig:mlhitop_general_2longest} 
\end{figure*}

Finally, the ML model is tested on TO designs that have multiple long members of nearly the same length.
Recall that the model is trained on ground truth annotations with \emph{one} human selection for each dataset topology.
Here, the human preference model for longest topological member demonstrates unexpected capabilities. 
Figure \ref{fig:mlhitop_general_2longest} presents two representative examples: the TO benchmark force inverter problem (Fig. \ref{fig:mlhitop_general_2longest}a) and a pin-pin beam problem (Fig. \ref{fig:mlhitop_general_2longest}e). 
The TO designs without human intervention for these two problems both contain two structural members whose lengths differ by less than 10\% from the maximum value (Fig. \ref{fig:mlhitop_general_2longest}b,f). 
Again, for visualization purposes, skeletonization is overlaid the TO designs in Fig. \ref{fig:mlhitop_general_2longest}d,h, where the two longest members are indicated with dashed lines.
As previously noted, the model has the ability to segment both prominent members, indicating that the model can holistically analyze topological features and identify structural patterns that extend beyond its explicit training objectives. 
The demonstrated capability to recognize and segment multiple critical members indicates advanced pattern recognition and represents a significant generalization beyond the original training dataset parameters.

\section{Conclusion}\label{sec:conclusion} 

This work introduces HiTopAI, a human-in-the-loop TO approach with AI co-pilot recommendations on where in the design to make changes.
The AI co-pilot is developed to mitigate the often time consuming trial-and-error bottleneck of manual intervention in existing human-in-the-loop TO design formulations. 
The recommendations are made by formulating the prediction of user-preferred modification regions as an image segmentation task.
The core ML model uses a U-Net architecture trained on synthetic data on human interventions.
The synthetic data generation initially identifies the longest topological feature as the only desired region for making design changes. 
The framework's performance is confirmed by showing that the AI-assisted process can achieve a 
39\% improvement in the buckling load performance of an L-bracket structure while minimally impacting the overall design time (a 15 sec or 4\% increase over conventional, simplistic TO). 
Furthermore, the model exhibits promising generalization capabilities, accurately segmenting critical features in diverse topologies.
It also notably develops the emergent ability to identify multiple near-maximum-length members despite being trained only on single-member annotations. 
This indicates that the model is learning deep structural patterns rather than merely mapping training labels.

The framework is also demonstrated on a separate dataset with a different human preference criterion that identifies the most complex topological connection.
Also here, the trained model presents satisfactory capability in classifying the node that connects most structural members.
Using the AI-assisted human-in-loop design approach with this data is shown to improve manufacturability by lowering the number of connected members and significantly increasing their angular spacing. 

A key insight put forth here is that because the AI co-pilot does not modify objectives or constraints, it cannot independently improve design performance. 
Conversely, human-only interaction requires iterative trial-and-error region selection. 
The presented results therefore establish a clear division of roles: the human provides domain judgment and decision authority, while the AI accelerates identification of intervention regions. 
The preliminary performance gains observed suggest that this collaboration has the potential to outperform either component alone.
We currently stipulate that it improves the overall TO workflow and time to achieve a design deemed \emph{acceptable} by the human engineer. However, future research that conducts comprehensive human-subject surveys is needed to empirically validate that the new framework can outperform individual experts or purely automated AI design in terms of design performance, time and workflow satisfaction.

This research establishes a foundation for the future integration of sophisticated ML co-agents that augment human expertise in high-performance engineering design.
For the model trained on the longest topological member, the average IOU of 0.58 validates the approach's proof-of-concept potential with acknowledged limitations.
One major limitation lies in the used human selection data. 
HiTopAI is trained on a synthetic dataset with imperfections in the automated skeletonization process used for ground-truth generation.
The reliance on skeletonization for training signals introduces potential noise where algorithmic labels may not perfectly align with human cognitive preferences. 
Moreover, the current evaluation focuses on pixel-level similarity rather than direct measurements of user iteration efficiency or design quality preservation. 
Moving forward, a crucial next step is to conduct user studies to create authentic datasets of human interaction patterns. This will allow the ML model to learn more nuanced preference criteria beyond the simple geometric features used in this study. 
Beyond dataset creation, future user studies should also provide the necessary empirical data to compare the integrated workflow's performance against traditional design, standard and human-in-the-loop TO in terms of design time, mechanical performance, manufacturabilty and design framework satisfaction. 
Future research should also expand the AI co-pilot's capabilities to incorporate multiple design criteria.
This will lead to a more comprehensive and versatile AI assistant in the inverse design workflow.

While fully autonomous discovery remains a long-term goal, this work argues that scalable engineering design will require systems that can recognize when their objectives are incomplete and efficiently collaborate with human experts to reshape the search space. 

\section{Methods}\label{sec:methods}

\subsection{Human-Informed Topology Optimization (HiTop)}\label{sec:to_hitop}

The first version of Human-Informed Topology Optimization (HiTop) \cite{2023_hitop} is used as the backbone of the current work. A brief description of how HiTop differs from classic TO is provided for completeness.

HiTop solves the most basic classic TO problem of minimum compliance, which seeks to find a stiff structure under static equilibrium with given loads and boundary conditions. 
Several approaches to solving classic TO problems have been presented in literature \cite{sigmund2013topology}. This work uses the so-called density-based approach \cite{1989_SIMP}. 
In density-based TO, the material distribution is represented by element density variables $\bar{\pmb{x}}$.
These quantities are connected through the finite elements within the design domain.
Each finite element $e$ is solid if $\bar{x}_e=1$ and void if $\bar{x}_e=0$.
Typically, the minimum compliance design problem is formulated as:
\begin{equation}
\begin{array}{ll}
    \underset{\pmb{x}}{\min}: &c(\pmb{x})=\pmb{U}^{T}\pmb{K}\pmb{U}\\
    \textrm{subject to:} & \pmb{K}\pmb{U}=\pmb{F} \\
    &V(\pmb{\bar{x}})/V_{0}=f  \\
   & 0 \le \pmb{x} \le 1 .
    \end{array}
\label{eq:88line}
\end{equation}
Here $\pmb{x}$ is the vector of design variables, bounded by $[0,1]$. 
These control the element densities $\bar{\pmb{x}}$.  
The compliance objective $c$ is calculated using the global displacement vector $\pmb{U}$ and the global stiffness matrix $\pmb{K}$. 
The first constraint states that the design must be in static equilibrium with the global force vector $\pmb{F}$ ($\pmb{K}\pmb{U}=\pmb{F}$).
Additionally, a volume fraction $f$ limits the total amount of material within the design domain. 
This constraint is expressed as the ratio of used material volume $V(\bar{\pmb{x}})$ to the volume of the design domain $V_{0}$. 
The compliance objective in Eq. (\ref{eq:88line}) is herein calculated as the sum of elemental compliances across all $N$ elements in the design domain:
\begin{equation}
c(\pmb{x})=\pmb{U}^{T}\pmb{K}\pmb{U}=\sum_{e=1}^{N}  E_{e}(\bar{x}_e)\pmb{u}_{e}^{T}\pmb{k}_{0e}\pmb{u}_{e},
\label{eq:c}
\end{equation}
where $\pmb{u}_{e}$ is the element displacement vector, and $\pmb{k}_{0e}$ is the stiffness matrix of a solid element. 

To enable use of gradient-based optimization to solve the problem in Eq. (\ref{eq:88line}), the design variables $\pmb{x}$ and the element densities $\bar{\pmb{x}}$ are left as continuous variables. 
To encourage a near-discrete final material distribution, density-based methods employ a penalization scheme. 
This work uses the Solid Isotropic Material Penalization (SIMP) method \cite{1989_SIMP,1992_SIMP}, where the stiffness of an element \(E_{e}(\bar{x}_e)\) is defined as:
\begin{equation}
    E_{e}(\bar{x}_{e})=E_{min}+\bar{x}_{e}^p(E_{0}-E_{min}). \label{eq:SIMP}
\end{equation}
Here, \(p\) denotes the SIMP penalty exponent, and \(E_{0}\) represents the solid material stiffness. 
A minimal stiffness \(E_{min}\) prevents numerical instabilities in void regions. 
All examples in this work use $p=3$, \(E_0 = 1\) and \(E_{min} = 10^{-9}\).

To prevent numerical instabilities such as checkerboarding, a direct link between design variables \(\pmb{x}\) and physical densities \(\bar{\pmb{x}}\) is avoided \cite{1995_diaz,1998_sigmund,2001_borrvall}. 
Instead, their relationship is defined through a filtering scheme. 
The filter operates within a neighborhood \(N_e\) of elements near element \(e\), comprising all elements \(i\) whose centroids \(\textbf{x}_i\) fall within a specified radius \(r_{min}\):  

\begin{equation}  
N_{e} = \{i \ \vert \ \| \textbf{x}_{i} - \textbf{x}_{e} \| \le r_{min} \}. \label{eq:Ne}  
\end{equation}

A linear density filter computes element densities by averaging design variables within the neighborhood \(N_e\) \cite{2001_bruns,2001_bourdin}:
\begin{equation}
    \tilde{x}_{e}=\frac{1}{\sum_{1 \in N_{e}} H_{ei}} \sum_{1 \in N_{e}} H_{ei}x_{i}. \label{eq:dens}
\end{equation}
Here, \(H_{ei} = \max(0,r_{min}-\Delta(e,i))\) represents the weight factor between elements \(e\) and \(i\).

This work further applies a Heaviside projection to ensure crisps topological boundaries \cite{2004_guest,2010_xu,2011_wang}.
Since all elements within the $N_e$ become solid when one $x_i=1$ and Heaviside projection is applied, it implicitly enforces a minimum size control on the topological features of the design.
The averaged variables \(\pmb{\tilde{x}}\) are processed through a regularized Heaviside function, where this work takes the formulation as in \cite{2010_xu,2011_wang}:
\begin{equation}
    \bar{x}_{e}=\frac{\tanh{\beta\eta}+\tanh{\beta(\tilde{x}_{e}-\eta)}}{\tanh{\beta\eta}+\tanh{\beta(1-\eta)}}. \label{eq:HPM_250}
\end{equation}
The parameter \(\beta\) controls approximation sharpness of the Heaviside function and \(\eta\) is a thresholding parameter which projects \(\tilde{x}_e > \eta\) to 1 and \(\tilde{x}_e < \eta\) to 0.
These are taken as 25 and 0.5 in all examples in this work.

To solve Eq. (\ref{eq:88line}) with gradient-based optimization, sensitivity analyses of the objective and constraint functions are needed. 
The standard sensitivity derivations are excluded here for conciseness and the reader is referred to  educational resources on density-based TO \cite{99line,88line,neo99line}. 
All examples in the current work uses the most common gradient-based optimizer for TO, namely, the
Method of Moving Asymptotes \cite{1987_MMA}.  




Classic TO takes a fully automated design approach and solves the design problem in Eq. (\ref{eq:88line}) till full convergence a single time. 
HiTop differs from classic TO by implementing a three-step process:
\begin{itemize}
\item[]  \emph{Step 1:} Initial compliance minimization solving Eq. (\ref{eq:88line}) for a fixed small number of iterations with a uniform minimum feature size ($r_{min}$) requirement.
\item[] \emph{Step 2:} User evaluation of the design quality, with optional interactive local feature size ($r_{min}$) adjustments in selected regions of the design.
\item[] \emph{Step 3:} Re-optimization to solve Equation (\ref{eq:88line}) using the modified feature size ($r_{min}$) map.
\end{itemize}
\emph{Steps 2--3} are repeated until the designer deems the result satisfactory.

The initial compliance minimization in \emph{Step 1} enables HiTop to take advantage of the exploratory power of automated TO. In this work, we limit the initial number of iterations in \emph{Step 1} to 50 as in the original paper \cite{2023_hitop}.
In \emph{Step 2}, the user can enhance the automated process by incorporating their expertise and identify and mark problematic design features while still letting the automated design framework suggest how to resolve these issues. 
For example, when selecting an region containing thin structural members and assigning a higher feature size ($r_{min}$) value, the algorithm can respond in \emph{Step 3} in several ways.
It can either reinforce the members in area with additional material, eliminate the problematic members entirely or combine to fewer, thicker members as in Fig. \ref{fig:hitop_MBB}.

\subsection{Synthetic Dataset Creation of Human Preference Selections}\label{sec:data_creation}

\begin{figure*}[h!]
    \centering
    \begin{tabular}{c}
    \includegraphics[scale = 0.8]{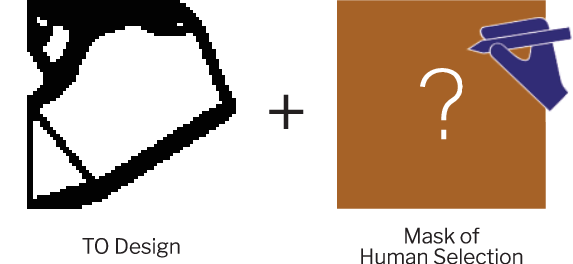} \\
    (a) \\[5 pt]
    \includegraphics[scale = 0.8]{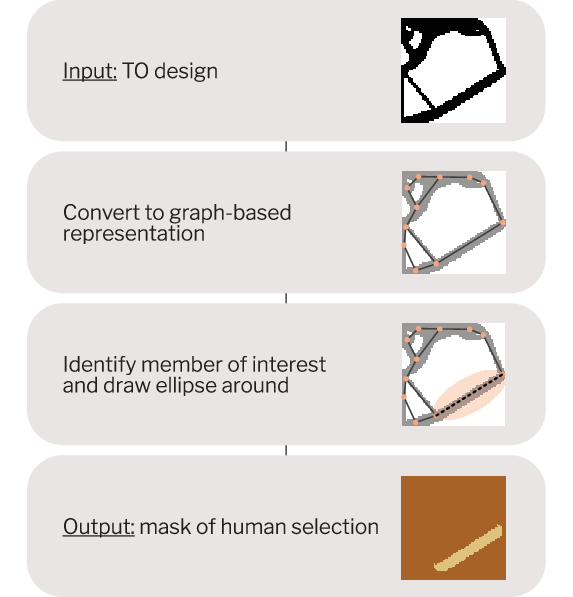} \\
    (b)
    \end{tabular}
    \caption{Schematic illustration of the synthetic dataset creation for the ML model on human preferences for regions to modify in TO designs. Training the model requires pairs with two data types; (a) a TO design and an image containing the corresponding mask of the human selection. Whereas datasets with TO designs are publicly available, there is no existing data on human preference selections. 
    This work generates masks with human selections using the process in (b). For each TO design in the dataset, skeletonization is applied to transform the topology to a graph-based representation, from which a topological feature of interest can be identified and enclosed in an ellipse. 
    The synthetic data generation scheme overlays the ellipse with the underlying topology to 
    create the mask of the human selection. }
    \label{fig:datasetcreation_schematic} 
\end{figure*}

As discussed, the ML model for human preference prediction is, at its core, an image segmentation model.
Therefore, training the ML model requires two types of data; TO designs with a corresponding segmentation mask.
As illustrated in Fig. \ref{fig:datasetcreation_schematic}a, there exists several public datasets with TO designs \cite{2017_sosnovik_todataset_NN12k,2023_topodiff,2024_li_todataset_doubleunet}, but currently there is no data available on human preference selections or the segmentation masks these will create. 
As will be discussed in the following, here we have used an automated process to synthetically generate the corresponding segmentation masks for an existing dataset of TO designs.
The process is schematically shown in Fig. \ref{fig:datasetcreation_schematic}b.
For each TO design within the dataset, skeletonization is performed to transform the TO design to a graph representation that reveals the underlying structure. Then, based on a specific criterion (in this paper, the longest topological member or most complex node), one particular topological feature is selected. Surrounding this feature, an ellipse is drawn, akin to the human selections in the original HiTop framework. The active material of the TO design that falls within this ellipse becomes the corresponding mask.

\subsubsection{Continuum TO dataset} \label{sec:topodiff}


While there have been multiple TO studies using ML, not all datasets are readily available \cite{2021_wang_todataset_deepgeneral, 2022_hoang_todataset_geometry, 2020_abueidda_todataset_2Dnonlinearity, 2020_kallioras_todataset_TOdeeplearn, 2020_lee_todataset_cnnimageTO}. 
Only a few public TO datasets exist.
These include 400,000 2D TO designs covering four boundary conditions at two resolutions \cite{2024_li_todataset_doubleunet}, 
and 10,000 sampled load nodes and fixed displacements on a 40 $\times$ 40 unit$^2$ mesh, with volume fractions drawn from a Gaussian distribution 
\cite{2017_sosnovik_todataset_NN12k}. 
This work starts with the dataset for the TopoDiff framework \cite{2023_topodiff}, which can be viewed as an improved version of the latter in terms of size and quality.
It contains 30,000 TO 2D designs, generated on a design domain with a 1:1 aspect ratio using a $64 \times 64$ unit$^2$ mesh. 
Within the TopoDiff dataset \cite{2023_topodiff}, each TO design is obtained by sampling the volume fraction uniformly in the interval $[0.3,0.5]$ with a step size of $0.02$. 
Boundary conditions are selected from 42 predefined scenarios, and loads are applied to randomly chosen unconstrained nodes along the domain’s boundary. The load direction is sampled in the interval 
$\pi$ at steps of $\frac{\pi}{6}$.
From our experimentation, we find it necessary to upscale the topologies from the original $64 \times 64$ unit$^2$ mesh to $128 \times 128$ unit$^2$ mesh to achieve stable performance in the skeletonization step described in Section \ref{sec:skeletonization}.


\subsubsection{Skeletonization} \label{sec:skeletonization}
\begin{figure*}[h!]
    \centering
    \begin{tabular}{cc}
    \includegraphics[scale = 0.8]{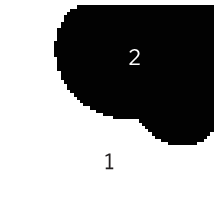} & 
    \includegraphics[scale = 0.8]{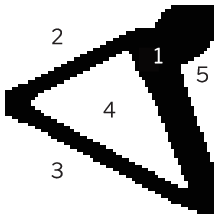} \\
    (a) & (b) \\[5 pt] 
    \includegraphics[scale = 0.8]{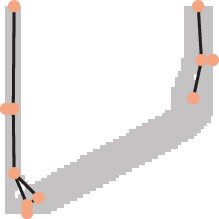} & 
    \includegraphics[scale = 0.8]{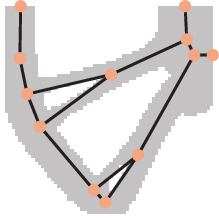} \\
    (c) & (d)
    \end{tabular}
    \caption{The existing dataset \cite{2023_topodiff} with 30,000 TO designs is filtered before it is used to generate synthetic human preference selection masks. First, TO designs that lack distinct topological features counted as the number of distinct image regions are removed and thereafter the TO designs are transformed to graph-based representations where incomplete graphs are filtered out. 
    The difference between TO designs with 2 and 5 distinct regions is shown in (a) and (b), where as the difference between incomplete and complete graph representations are given in (c) and (d). }
    \label{fig:DataFilter} 
\end{figure*}

As mentioned, this work synthetically generates human preference selections
instead of making manual selections on the 30,000 TO designs within the existing dataset.
However, prior to generating the synthetic human selections, a two-step filtering process is applied to the $128 \times 128$ unit$^2$ mesh upscaled TopoDiff training dataset.
First, topologies lacking distinct structural features are filtered out using MATLAB’s image segmentation tool \cite{matlab_doc}. 
The topology image is partitioned into distinct regions. TO designs with fewer than three segments are discarded as the low number of distinct regions indicate poorly defined features.
Two example topologies are shown in Fig. \ref{fig:DataFilter}a and b, where 
Fig. \ref{fig:DataFilter}a is a topology with 2 distinct regions whereas the TO design in Fig. \ref{fig:DataFilter}b has 5 distinct regions. 
Filtering out TO designs with $2\leq$ distinct features removes approximately 6,000 data points, leaving 24,000 TO designs with identifiable structural members.
Next, the remaining TO designs undergo skeletonization based on the method by Xia et al. \cite{2020_xia_autostruttie, 2020_xia_struttieconcrete}, which converts continuum material distributions into graph-based truss representations. 
This process identifies nodes and their connectivity, enabling extraction of different kinds of topological features such as member lengths and thickness or node complexity. 
However, the used skeletonization algorithm occasionally produces incomplete graphs. 
When this occurs, the graph-based representation of the structure is disconnected and key structural members are not captured. 
The difference between incomplete and complete graph representations can be seen by comparing the two skeletonized TO design in Fig. \ref{fig:DataFilter}c and d.

To lower the number of discontinuous skeletons, some modifications to the original skeletonization algorithm are performed.
The original skeletonization method requires manual identification of seeding points to align with problem-specific definitions such as applied loads and boundary conditions. 
This manual definition becomes impractical when applied to large-scale datasets. 
It is particularly ineffective when entire edges are constrained, as the algorithm cannot automatically remove fixed seeding points.
To address this limitation, this work automates the seeding point selection by analyzing the final material distribution. 
Through empirical observation, it is found that robust skeletonization depends on accurately identifying boundary nodes where material contacts the design domain edges. 
For material strips along these edges, seeding points are placed at their midpoints, eliminating the need for manual intervention while preserving geometric accuracy.

With the automated seeding points in place, the original skeletonization algorithm still frequently produces incomplete truss representations, omitting critical nodal connections that are structurally significant.
Even with this modification, the skeletonization developed by Xia et al. \cite{2020_xia_autostruttie, 2020_xia_struttieconcrete} only yields a fully connected graph for 58.3\% of the input topologies in our dataset. To significantly improve graph continuity and accuracy, we employ additional topology-specific enhancements to the skeletonization method.
The specific of the modifications are listed below. 
When implemented, we observe an improvement in the number of fully connected graphs from 58.3\%  to 76.0\%.

The first crucial modification addresses the cell partitioning and boundary handling of the design space. 
In the original framework, the design space is divided based on a 1-pixel thick skeleton created by thinning the solid topology. 
However, the original segmentation process fails to delineate distinct sections when voids border the exterior of the design domain. 
This impacts the subsequent connectivity step, 
and frequently results in discontinuous graphs. 
Consequently, the initial framework is modified to ensure that the segmentation accurately defines and includes all cells bordered by both the skeleton and the design boundary.

Second, once all sections have been correctly partitioned, the process of determining node connectivity is refined through a guaranteed traversal algorithm. 
The existing method establishes connectivity by walking through the thinning result within the bounds of a given section.
We further refine this to ensure that the entire portion of the skeleton must be traversed, leaving no potential node unchecked or unconnected. 
If the section is bounded by the design domain, the starting point for the traverse must be a boundary node, whereas an internal section must successfully form a closed loop upon completion. 
When traversing the thinning section, the method is modified such that at a given position, the algorithm first checks the cardinal directions to identify nearby unexplored pixels, and the position is updated to the first one found. 
The pixels in the diagonal positions are explored if there are no nearby unexplored pixels in the cardinal directions. 
Each time a pixel is traversed, it is marked as explored. The traverse algorithm is completed when the counter of explored pixels is equal to the total number of pixels in the segment, guaranteeing exhaustive discovery.

Finally, once complete node connectivity has been established, a topology-aware node refinement process is used. 
Here, nodes that are within a 15-pixel radius of each other are identified as candidates for merging. 
A selection scheme is then proposed where one node is identified as the incoming node (which will be merged), and the other is termed the host node (which retains its position). 
When the incoming node, which is connected to other distant nodes, is merged, the host node must carry over the incoming node’s structural members. 
For each of these new potential merged connections, a metric is devised to calculate the portion of a straight line graphing the topology drawn that remains within the solid material of the continuum representation. 
Based on this metric, the host node is chosen as the node that maximizes this metric.

As mentioned, these modifications have a significant impact on the success rate of the skeletonization step. 
Nevertheless, unconnected graphs still occasionally occur (Fig. \ref{fig:DataFilter}c). 
In our experience, incomplete connectivities tend to arise when the input topologies are either inherently geometrically challenging or contain imperfections such as floating material or local checkerboarding.
A graph connectivity test is applied to detect and discarding any topologies with graphs that are not fully connected.
Approximately 6000 TO structures fail the graph connectivity test, leaving 24,000 TO designs with complete graph representations such as the example in Fig. \ref{fig:DataFilter}d. 

\subsubsection{Selecting member of interest and mask creation} \label{sec:maskcreation}

The graph representation of each TO design is used to enable an automated mask generation that represents the synthetic human selection of a region where modifications are desired. 
From the graph representation, it is possible to identify members within the design with different characteristics such as the longest, shortest, thinnest, etc.
In this work, the synthetic human selections define the longest structural member and most complex node within a TO design as this is visually the easiest to communicate. 
However, the procedure used herein can be employed to generate synthetic datasets with various other structural characteristics.

\begin{figure}
\centering
\begin{tabular}{c}
\includegraphics[width = 0.9\linewidth]{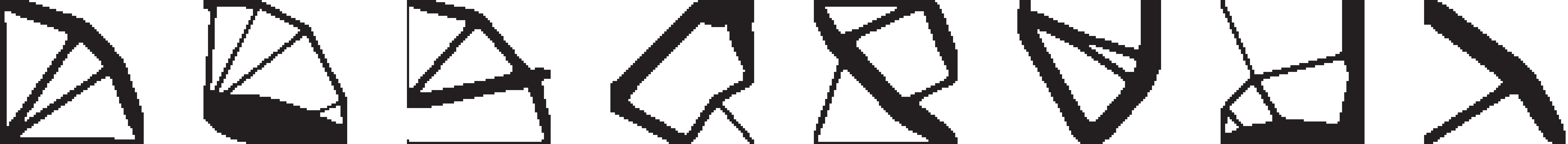} \\
(a) \\[5 pt]
\includegraphics[width = 0.9\linewidth]{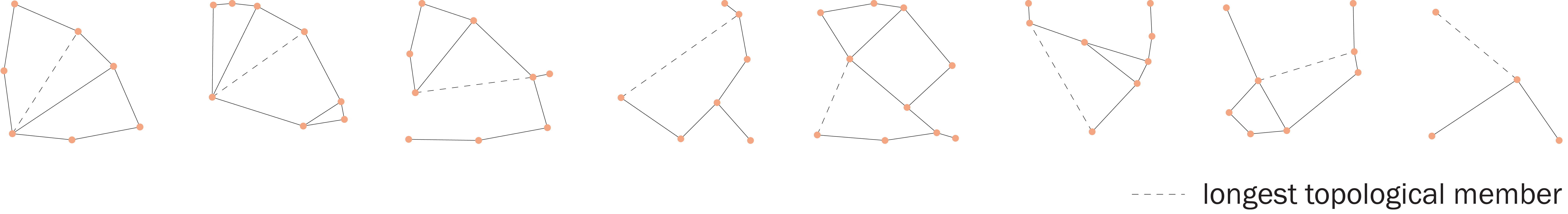} \\
(b) \\[5 pt]
\includegraphics[width = 0.9\linewidth]{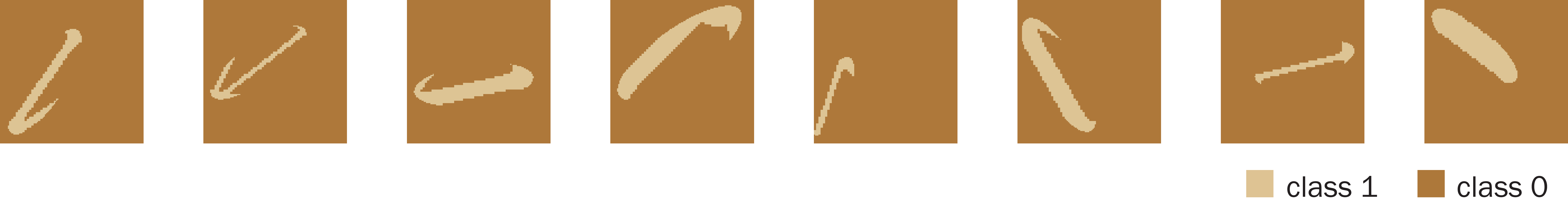} \\
(c) \\[5 pt]
\includegraphics[width = 0.9\linewidth]{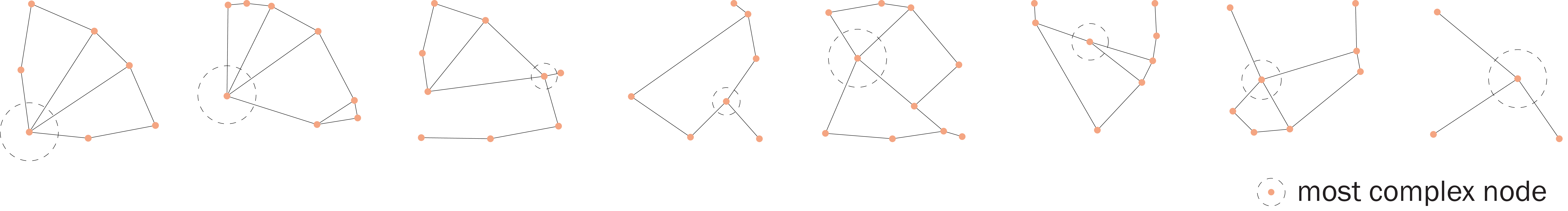} \\
(d) \\[5 pt]
\includegraphics[width = 0.9\linewidth]{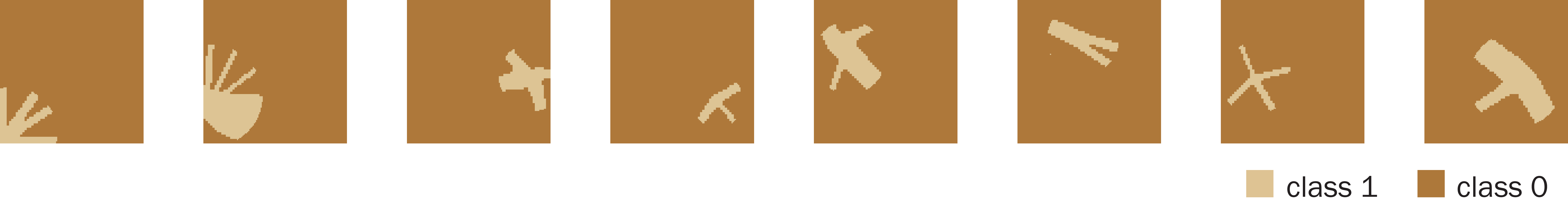} \\
(e) \\[5 pt]
\end{tabular}
    \caption{Examples of eight random (a) designs from the filtered TopoDiff dataset \cite{2023_topodiff}. (b) The TO designs undergo skeletonization to represent the structures as graphs. For the longest member dataset, the longest edge is identified. (c) An elliptical region is drawn around the longest member and the intersection of pixels within the ellipse and solid pixels in the TO design form the segmentation mask that describes the synthetically generated human selection for regions to change (class 1). (d) For the most complex node, the node with the highest connectivity degree is identified. (e) A circular region is drawn around this node, with the radius taken as the average length of the edge members connected to this node.}
    \label{fig:TopoMask_Dataset}
\end{figure}

When selecting the longest member, the longest continuous edge in the graph representation is identified for each TO design in filtered the dataset, 
This becomes the major axis of an elliptical region of interest. 
For proportional scaling, the minor axis is set to one-third of this length.
Pixels inside the ellipse are labeled class 1, marking user-specified regions. 
All others receive class 0, designating unselected areas. 
Then, we take the intersection of the pixels in the elliptical region and the solid part of the TO design to form the binary segmentation mask.
Our experimentation shows that classifying only the solid pixels within the TO design's member of interest allows for a cleaner segmentation task definition.
The process for defining the region of the most complex node also starts with the skeletonization of the designs.
From the graph representation, the node that connects the highest number of structural members is identified.
Surrounding this node, a circular region is drawn.
The radius of this region is calculated as the average length of the edges connected to the node.
Examples of some of the dataset's TO designs and their corresponding selection masks are shown in Fig. \ref{fig:TopoMask_Dataset}, indicating that the synthetic human preference selections works across diverse structures.

To expand the dataset, the remaining topologies are augmented through geometric transformations. 
Each topology is rotated counter-clockwise by 90°, 180°, and 270° and mirrored horizontally and vertically, yielding six variants per sample. 
This augmentation increases the final dataset size to 109,722 topology-mask pairs.
The used data generation and augmentation pipeline ensures that ML human preference model is trained on high-quality, diverse data, with explicit segmentation masks derived from meaningful structural features. 

\subsection{Machine Learning Model and Post-Processing}\label{sec:mlhitop_model_postprocess}

\subsubsection{Machine Learning model}



\begin{figure}[h!]
    \centering
    \includegraphics[width = \linewidth]{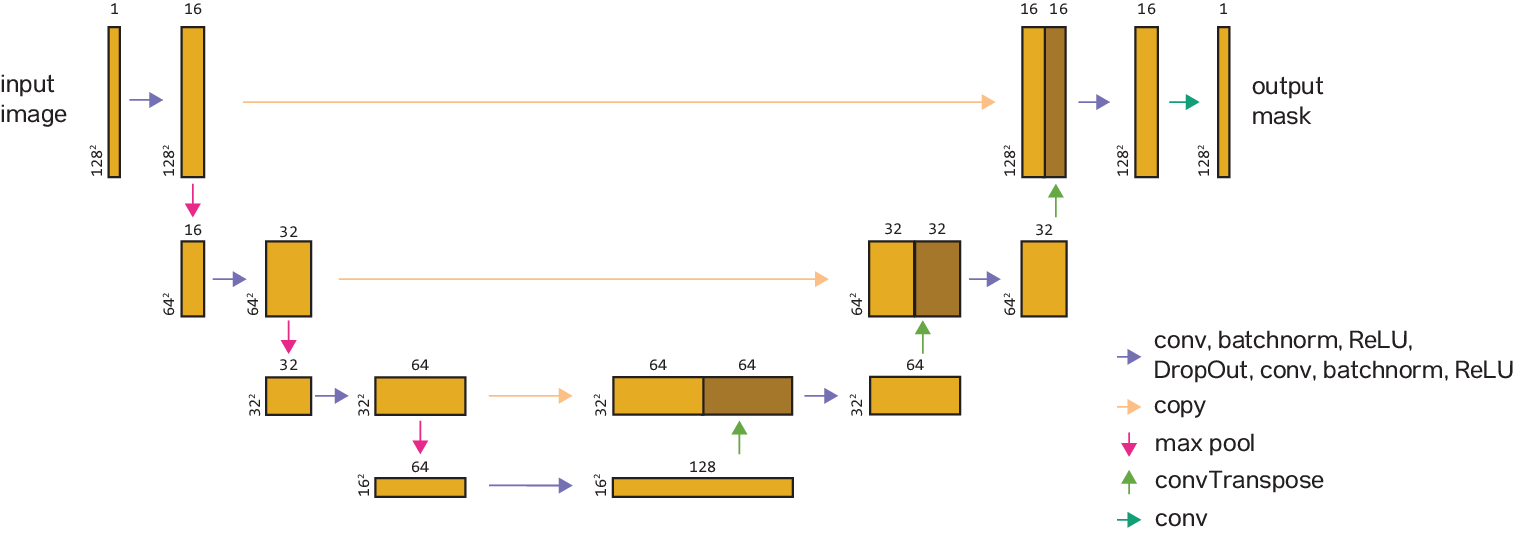}
    \caption{The machine learning model uses a U-Net architecture that is adapted for $128 \times 128$ resolution input images representing of TO designs. Each double convolution block (purple arrows) incorporates dropout layers to enhance model robustness. The model processes input topologies and generates segmentation masks classifying pixels as either human selections (class 1) or outside the user selected region (class 0).}
    \label{fig:ML_Architecture}
\end{figure}

The ML model used in this work to model human preference selections is a U-Net architecture, specifically designed for the herein considered image segmentation task. 
The backbone of the architecture is inspired by the U-Net used to predict neuronal structures identified by medical experts \cite{2015_ronnebberger_unet}.
The network is illustrated in Fig. \ref{fig:ML_Architecture}.
It consists of an encoder path with three double convolutional layers and max pooling for downsampling, a bottleneck section, and a decoder path with three double convolutional layers and transposed convolutions for upsampling. 
The final layer is a 1$\times$1 convolution to produce a single-channel output. 
Dropout with a rate of 0.15 is included in the double convolutional layers to help prevent overfitting. 
The training scheme utilizes the BCEWithLogitsLoss as the criterion and the Adam optimizer with a learning rate of 0.0005. 
The model is trained for a maximum of 1000 epochs with an early stopping mechanism based on the test loss, stopping if the test loss does not improve by at least 0.01 for 25 consecutive epochs.
The total dataset of 109,722 pairs is partitioned into a training set (70\%), a validation set (10\%), and a true held-out test set (20\%). 
The model is trained using the training set, while the validation set is used to monitor performance and implement early stopping as the termination criterion. 
Final performance metrics are reported solely on the test set, which remains unseen by the system throughout the training and tuning phases.



\subsubsection{Post Processing}

Following the ML model's output, post-processing is applied to obtain the final binary segmentation mask. 
The model outputs a continuous prediction of the labels for the pixels in the image of the structural topology.
This output is passed through a sigmoid activation function, which in turn passes through a 90th percentile.
Pixels with output values greater than or equal to this percentile threshold are set to 1, while all others are set to 0. 
This ensures a binary segmentation mask.
An additional post-processing step is required to obtain the elliptical region that will be presented to the human user as the AI-co pilot's recommended region for making changes.
In the last post-processing step, the largest connected component of pixels in the thresholded mask is identified and the smallest ellipse is fitted around the pixel cluster.

\medskip
\textbf{Supporting Information} \par 
Supporting Information is available from the author.

\medskip
\textbf{Acknowledgments} \par 
The Funding Support from the MathWorks Engineering Fellowship Fund is gratefully acknowledged.

\medskip

\textbf{Author contributions} \par 
D. Q. Ha contributed the research and writing of the manuscript.
J. V. Carstensen contributed the funding, advice, writing and editing of the manuscript.
F. Alam, M. J. Buehler, and F. Ahmed contributed with advice during the research phase of the work.

\medskip

\textbf{Competing interests}
The authors declare no competing interests.

\medskip

\textbf{Materials \& Correspondence}
The corresponding author D. Q. Ha can be contacted through email dath@mit.edu for any correspondence and material requests.

%
\bibliographystyle{MSP}
\bibliography{_references}

@article{2023_hitop,
	author	=	"Dat Q. Ha and Josephine V. Carstensen",
	title	=	"Human‑Informed Topology Optimization: interactive application of feature size controls",
	journal =   "Structural and Multidisciplinary Optimization",
	volume  =	"66",
	number	=	"59",
	pages   =   "",
	month   =  "",
	year	=	2023}

@article{2024_gillian_infill,
	author	=	"Gillian Schiffer and Martin-Pierre Schmidt and Claus B. W. Pedersen and Josephine V. Carstensen",
	title	=	"Interactive infill topology optimisation guided by user drawn patterns",
	journal =   "Virtual and Physical Prototyping",
	volume  =	"19",
	number	=	"1",
	pages   =   "",
	month   =  "",
	year	=	2024}

@article{2023_gillian_hitop2,
	author	=	"Gillian Schiffer and Dat Quoc Ha and Josephine V. Carstensen",
	title	=	"HiTop 2.0: combining topology optimisation
with multiple feature size controls and human preferences",
	journal =   "Virtual and Physical Prototyping",
	volume  =	"19",
	number	=	"1",
	pages   =   "",
	month   =  "",
	year	=	2023}

@article{2026_tian_dpdnto,
	author	=	"Boxu Tian and Wenliang Qian and Yefei Yang",
	title	=	"Designer preference-driven topology optimization using a human-in-the-loop neural network",
	journal =   "Computers and Structures",
	volume  =	"321",
	number	=	"",
	pages   =   "",
	month   =  "",
	year	=	2026}

@article{2025_aragon,
	author	=	"Alejandro M. Aragón and Hendrik J. Algra",
	title	=	"ARCADE: An interactive playground for real-time immersed topology optimization",
	journal =   "arXiv: Computational Engineering, Finance, and Science",
	volume  =	"1",
	number	=	"2501.13564",
	pages   =   "",
	month   =  "",
	year	=	2025}

@article{2024_zhang,
	author	=	"Weisheng Zhang and Yue Wang and Sung-Kie Youn and Xu Guo",
	title	=	"Machine learning powered sketch aided design via
topology optimization",
	journal =   "Computer Methods in Applied Mechanics
and Engineering",
	volume  =	"1",
	number	=	"419",
	pages   =   "",
	month   =  "",
	year	=	2024}

@article{2021_garrelts,
	author	=	"Enno Garrelts and Marco Huberc and Daniel Rothb and Hansgeorg Binzb",
	title	=	"AI-Based Topology Optimization of Freehand Sketches",
	journal =   "Procedia CIRP",
	volume  =	"104",
	number	=	"419",
	pages   =   "1316-1321",
	month   =  "",
	year	=	2021}

@article{2025_zhu,
	author	=	"Siyu Zhu and Jie Hu and Jin Qi and Lingyu Wang and Jing Guo and Jin Ma and Guoniu Zhu",
	title	=	"Sketch-Guided Topology Optimization with Enhanced Diversity for Innovative Structural Design",
	journal =   "Applied Sciences",
	volume  =	"15",
	number	=	"2753",
	pages   =   "",
	month   =  "",
	year	=	2025}

@article{2023_topodiff,
	author	=	"Fran\c{c}ois Maz\'{e} and Faez Ahmed",
	title	=	"Diffusion models beat GANs on topology optimization",
	journal =   "AAAI Press",
	volume  =	"1",
	number	=	"1024",
	pages   =   "9",
	month   =  "",
	year	=	2023}

@article{2020_xia_autostruttie,
	author	=	"Yi Xia and Matthijs Langelaar and Max A.N. Hendriks",
	title	=	"Automated optimization-based generation and quantitative evaluation of Strut-and-Tie models",
	journal =   "Computers and Structures",
	volume  =	"238",
	number	=	"106297",
	pages   =   "",
	month   =  "",
	year	=	2020}

@article{2020_xia_struttieconcrete,
	author	=	"Yi Xia and  Matthijs Langelaar and  Max A.N. Hendriks",
	title	=	"A critical evaluation of topology optimization results for strut-and-tie modeling of reinforced concrete",
	journal =   "Comput Aided Civil and Infrastructure Engineering",
	volume  =	"35",
	number	=	"",
	pages   =   "850-869",
	month   =  "",
	year	=	2020}

@article{2016_zhu_aircraft,
	author	=	"Ji-Hong Zhu and Wei-Hong Zhang and Liang Xia ",
	title	=	"Topology Optimization in Aircraft and Aerospace Structures Design",
	journal =   "Arch Computat Methods Eng",
	volume  =	"23",
	number	=	"",
	pages   =   "595-622",
	month   =  "",
	year	=	2016}

@article{2011_tomlin_aircraft,
	author	=	"Matthew Tomlin and Jonathan Meyer",
	title	=	"Topology Optimization of an Additive Layer Manufactured (ALM) Aerospace Part",
	journal =   "7th Altair CAE Technology Conference",
	volume  =	"7",
	number	=	"",
	pages   =   "",
	month   =  "",
	year	=	2011}

@article{2021_zhang_dome,
	author	=	"Jingwei Zhang and Jun Yanagimoto",
	title	=	"Topology optimization of microlattice dome with enhanced stiffness and energy absorption for additive manufacturing",
	journal =   "Composite Structures",
	volume  =	"255",
	number	=	"112889",
	pages   =   "",
	month   =  "",
	year	=	2021}

@article{2020_yang_lattice,
	author	=	"Chengxing Yang and Qing Ming Li",
	title	=	"Advanced lattice material with high energy absorption based on topology optimisation",
	journal =   "Mechanics of Materials",
	volume  =	"148",
	number	=	"103536",
	pages   =   "",
	month   =  "",
	year	=	2020}

@article{2003_kiziltas_dielectric,
	author	=	"G. Kiziltas and D. Psychoudakis and J. L. Volakis and N. Kikuchi",
	title	=	"Topology design optimization of dielectric substrates for bandwidth improvement of a patch antenna",
	journal =   "IEEE Transactions on Antennas and Propagation",
	volume  =	"51",
	number	=	"10",
	pages   =   "2732-2743",
	month   =  "",
	year	=	2003}

@article{2021_wu_thermal,
	author	=	"Shuhao Wu and Yongcun Zhang and Shutian Liu",
	title	=	"Transient thermal dissipation efficiency based method for topology optimization of transient heat conduction structures",
	journal =   "International Journal of Heat and Mass Transfer",
	volume  =	"170",
	number	=	"121004",
	pages   =   "",
	month   =  "",
	year	=	2021}

@article{2020_pang_bellcrank,
	author	=	"Toh Yen Pang and Mohammad Fard",
	title	=	"Reverse Engineering and Topology Optimization for Weight-Reduction of a Bell-Crank",
	journal =   "Applied Sciences",
	volume  =	"10",
	number	=	"23",
	pages   =   "",
	month   =  "",
	year	=	2020}

@article{2025_zhang_MMCSIMP,
	author	=	"Weisheng Zhang and Xiaoyu Zhuang and Xu Guo and Xu Guo and Sung-Kie Youn",
	title	=	"Human-augmented topology optimization design with multi-framework intervention",
	journal =   "Engineering with Computers",
	volume  =	"",
	number	=	"",
	pages   =   "",
	month   =  "",
	year	=	2025}

@article{2019_li_photonic,
	author	=	"Weibai Li and Fei Meng and Yafeng Chen and Yang fan Li and Xiaodong Huang",
	title	=	"Topology Optimization of Photonic and Phononic Crystals and Metamaterials: A Review",
	journal =   "Advanced Theory and Simulations",
	volume  =	"2",
	number	=	"1900017",
	pages   =   "",
	month   =  "",
	year	=	2019}

@article{2015_ronnebberger_unet,
	author	=	"Olaf Ronneberger and Philipp Fischer and Thomas Brox",
	title	=	"U-Net: Convolutional Networks for Biomedical Image Segmentation",
	journal =   "Medical Image Computing and Computer-Assisted Intervention – MICCAI 2015",
	volume  =	"9351",
	number	=	"",
	pages   =   "",
	month   =  "",
	year	=	2015}

@article{alkhatib2023isotropic,
  title={Isotropic energy absorption of topology optimized lattice structure},
  author={Alkhatib, Sami E and Karrech, Ali and Sercombe, Timothy B},
  journal={Thin-Walled Structures},
  volume={182},
  pages={110220},
  year={2023},
  publisher={Elsevier}
}

@article{zeng2023inverse,
  title={Inverse Design of Energy-Absorbing Metamaterials by Topology Optimization},
  author={Zeng, Qingliang and Duan, Shengyu and Zhao, Zeang and Wang, Panding and Lei, Hongshuai},
  journal={Advanced Science},
  volume={10},
  number={4},
  pages={2204977},
  year={2023},
  publisher={Wiley Online Library}
}

@article{chen2018design,
  title={Design of buckling-induced mechanical metamaterials for energy absorption using topology optimization},
  author={Chen, Qi and Zhang, Xianmin and Zhu, Benliang},
  journal={Structural and Multidisciplinary Optimization},
  volume={58},
  number={4},
  pages={1395--1410},
  year={2018},
  publisher={Springer}
}

@article{carstensen2022topology,
  title={Topology-optimized bulk metallic glass cellular materials for energy absorption},
  author={Carstensen, Josephine V and Lotfi, Reza and Chen, Wen and Szyniszewski, Stefan and Gaitanaros, Stavros and Schroers, Jan and Guest, James K},
  journal={Scripta Materialia},
  volume={208},
  pages={114361},
  year={2022},
  publisher={Elsevier}
}

@article{dalklint2023computational,
  title={Computational design of metamaterials with self contact},
  author={Dalklint, Anna and Sj{\"o}vall, Filip and Wallin, Mathias and Watts, Seth and Tortorelli, Daniel},
  journal={Computer Methods in Applied Mechanics and Engineering},
  volume={417},
  pages={116424},
  year={2023},
  publisher={Elsevier}
}

@inproceedings{liu2020experimental,
  title={Experimental investigation of topology-optimized deep reinforced concrete beams with reduced concrete volume},
  author={Liu, Yan and Jewett, Jackson L and Carstensen, Josephine V},
  booktitle={RILEM International Conference on Concrete and Digital Fabrication},
  pages={601--611},
  year={2020},
  organization={Springer}
}

@article{pressmair2023contribution,
  title={A contribution to resource-efficient construction: design flow and experimental investigation of structurally optimised concrete girders},
  author={Pressmair, Nadine and Kromoser, Benjamin},
  journal={Engineering Structures},
  volume={281},
  pages={115757},
  year={2023},
  publisher={Elsevier}
}

@article{wethyavivorn2022topology,
  title={Topology optimization-based reinforced concrete beams: design and experiment},
  author={Wethyavivorn, Benjapon and Surit, Siradech and Thanadirek, Thanit and Wethyavivorn, Piyanut},
  journal={Journal of Structural Engineering},
  volume={148},
  number={10},
  pages={04022154},
  year={2022},
  publisher={American Society of Civil Engineers}
}

@inproceedings{schiffer2024integrating,
    author = {Schiffer, G. L. and Ha, D. Q. and Carstensen, J. V.},
    title = {Integrating Topology Optimization and Aesthetic Design Preferences through Interactive User-Sketched Input},
    booktitle = {Proceedings of the IASS 2024 Symposium - {Redefining} the Art of Structural Design},
    editor = {Block, Philippe and Boller, Giulia and DeWolf, Catherine and Pauli, Jacqueline and Kaufmann, Walter},
    pages={nill},
    year={2024}
}

@article{2025_li_vrbeso,
  title={Interactive 3D structural design in virtual reality using preference-based topology optimization},
  author={Zhi Li and Ting-Uei Lee and Yi Min Xie},
  journal={Computer-Aided Design},
  volume={180},
  pages={103826},
  year={2025},
  publisher={Elsevier}
}

@article{2023_li_besoscoring,
  title={Interactive Structural Topology Optimization with Subjective Scoring and Drawing Systems},
  author={Zhi Li and Ting-Uei Lee and Yi Min Xie},
  journal={Computer-Aided Design},
  volume={160},
  pages={103532},
  year={2023},
  publisher={Elsevier}
}

@article{2024_li_todataset_doubleunet,
  title={Topology optimization based on improved DoubleU-Net using four boundary condition datasets},
  author={Shun Li and PeiJian Zeng and Nankai Lin and Maohua Lu and Jianghao Lin and Aimin Yang},
  journal={Engineering Optimization},
  volume={57},
  pages={884-902},
  year={2024},
  publisher={Taylor & Francis}
}

@article{2021_wang_todataset_deepgeneral,
  title={A deep convolutional neural network for topology optimization with perceptible generalization ability},
  author={Dalei Wang and Cheng Xiang and Yue Pan and Airong Chen and Xiaoyi Zhou and Yiquan Zhang},
  journal={Engineering Optimization},
  volume={11},
  pages={1-16},
  year={2021},
  publisher={Taylor & Francis}
}

@article{2022_hoang_todataset_geometry,
  title={Data-driven geometry-based topology optimization},
  author={Van-Nam Hoang and Ngoc-Linh Nguyen and Dat Q. Tran and Quang-Viet Vu and H. Nguyen-Xuan},
  journal={Structural and Multidisciplinary Optimization},
  volume={65},
  number={69},
  year={2022},
  publisher={Springer}
}

@article{2020_abueidda_todataset_2Dnonlinearity,
  title={Topology optimization of 2D structures with nonlinearities using deep learning},
  author={Diab W. Abueidda and Seid Koric and Nahil A. Sobh},
  journal={Computers and Structures},
  volume={237},
  year={2020},
  publisher={Elsevier}
}

@article{2017_sosnovik_todataset_NN12k,
  title={Neural networks for topology optimization},
  author={Ivan Sosnovik and Ivan Oseledets},
  journal={Russian Journal of Numerical Analysis and Mathematical Modelling},
  volume={34},
  number ={4},
  year={2017},
  pages={215-223},
  publisher={De Gruyter}
}

@article{2020_kallioras_todataset_TOdeeplearn,
  title={Accelerated topology optimization by means of deep learning},
  author={Nikos Ath. Kallioras and Georgios Kazakis and Nikos D. Lagaros},
  journal={Structural and Multidisciplinary Optimization},
  volume={62},
  year={2020},
  pages={1185–1212},
  publisher={Springer}
}

@article{2020_lee_todataset_cnnimageTO,
  title={CNN-based image recognition for topology optimization},
  author={Seunghye Lee and Hyunjoo Kim and Qui X. Lieu and Jaehong Lee},
  journal={Knowledge-Based Systems},
  volume={198},
  year={2020},
  number={105887},
  publisher={Elsevier}
}

@article{2016_nobeljorgensen_togame,
  title={Improving topology optimization intuition through games},
  author={Morten Nobel-Jorgensen and  David Malmgren-Hansen and J. Andreas Bærentzen and Ole Sigmund and Niels Aage},
  journal={Structural and Multidisciplinary Optimization},
  volume={54},
  year={2016},
  number={4},
  pages={775–781},
  publisher={Springer}
}

@article{1998_sigmund,
	author	=	"Ole Sigmund and Joakim Petersson",
	title	=	"Numerical Instabilities in Topology Optimization: A Survey on Procedures Dealing with Checkerboards, Mesh-Dependencies and Local Minima",
	journal =   "Structural and Multidisciplinary Optimization",
	volume  =	16,
	number	=	"1",
	pages   =   "68--75",
	year	=	1998}

@article{2023_shin,
	author	=	"Seungyeon Shin and Dongju Shin and Namwoo Kang",
	title	=	"Topology optimization via machine learning and deep learning: a review ",
	journal =   "Journal of Computational Design and Engineering",
	volume  =	10,
	number	=	4,
	pages   =   "1736-1766",
	month   =   "",
	year	=	2023}

@article{1987_MMA,
	author	=	"K. Svanberg",
	title	=	"The method of moving asymptotes: A new method for structural optimization",
	journal =   "International journal for numerical methods in engineering",
	volume  =	24,
	number	=	2,
	pages   =   "359--373",
	month   =  "",
	year	=	1987}

@article{88line,
  title={Efficient topology optimization in MATLAB using 88 lines of code},
  author={Andreassen, Erik and Clausen, Anders and Schevenels, Mattias and Lazarov, Boyan S and Sigmund, Ole},
  journal={Structural and Multidisciplinary Optimization},
  volume={43},
  number={1},
  pages={1--16},
  year={2011},
  publisher={Springer}
}

@book{matlab_doc, 
year = {2022}, 
author = {MATLAB}, 
title = {version (2022a)}, 
publisher = {The MathWorks Inc.}, 
address = {Natick, Massachusetts} 
}

@article{1995_diaz,
	author	=	"Alejandro Diaz and Ole Sigmund",
	title	=	"Checkerboard Patterns in Layout Optimization",
	journal =   "Structural and Multidisciplinary Optimization",
	volume  =	10,
	number	=	1,
	pages   =   "40--45",
	month   =  "",
	year	=	1995}

@article{2001_borrvall,
	author	=	"Thomas Borrvall",
	title	=	"Topology Optimization of Elastic Continua using Restriction",
	journal =   "Archives of Computational Methods in Engineering",
	volume  =	8,
	number	=	"4",
	pages   =   "351--385",
	year	=	2001}

@article{2010_xu,
	author	=	"Shengli Xu and Yuanwu Cai and Gengdong Cheng",
	title	=	"Volume Preserving Nonlinear Density Filter Based on Heaviside Functions",
	journal =   "Structural and Multidisciplinary Optimization",
	volume  =	41,
	number	=	"4",
	pages   =   "495--505",
	month   =  "",
	year	=	2010}

@article{2011_wang,
	author	=	"Fengwen Wang and Boyan S. Lazarov and Ole Sigmund",
	title	=	"On Projection Methods, Convergence and Robust Formulations in Topology Optimization",
	journal =   "Structural and Multidisciplinary Optimization",
	volume  =	43,
	number	=	"6",
	pages   =   "767--784",
	month   =  "",
	year	=	2011}

@article{2001_bourdin,
	author	=	"B. Bourdin",
	title	=	"Filters in topology optimization",
	journal =   "International Journal for Numerical Methods in Engineering",
	volume  =	50,
	number	=	"9",
	pages   =   "2143--2158",
	month   =  "",
	year	=	2001}

@article{2001_bruns,
	author	=	"T. E. Bruns and D. A. Tortorelli",
	title	=	"Topology optimization of non-linear elastic structures and compliant mechanisms",
	journal =   "Computer Methods in Applied Mechanics and Engineering",
	volume  =	190,
	number	=	"26",
	pages   =   "3443--3459",
	month   =  "",
	year	=	2001}

@article{2004_guest,
	author	=	"J. K. Guest and J. H. Pr\'{e}vost and T. Belytschko",
	title	=	"Achieving minimum length scale in topology optimization using nodal design variables and projection functions",
	journal =   "International Journal for Numerical Methods in Engineering",
	volume  =	61,
	number	=	2,
	pages   =   "238--254",
	month   =  "",
	year	=	2004}

@article{99line,
	author	=	"Ole Sigmund",
	title	=	"A 99 line topology optimization code written in Matlab",
	journal =   "Structural and Multidisciplinary Optimization",
	volume  =	21,
	number	=	"",
	pages   =   "120-127",
	month   =  "",
	year	=	2001}

@article{neo99line,
	author	=	"Federico Ferrari and Ole Sigmund",
	title	=	"A new generation 99 line Matlab code for compliance Topology Optimization and its extension to 3D",
	journal =   "Structural and Multidisciplinary Optimization",
	volume  =	62,
	number	=	"",
	pages   =   "2211-2228",
	month   =  "",
	year	=	2020}

@article{2022_woldseth,
	author	=	"Rebekka V. Woldseth and Niels Aage and J. Andreas B{\ae}rentzen and Ole Sigmund",
	title	=	"On the use of Artificial Neural Networks in Topology Optimisation",
	journal =   "Structural and Multidisciplinary Optimization",
	volume  =	65,
	number	=	10,
	pages   =   294,
	month   =  "",
	year	=	2022}

@article{1989_SIMP,
	author	=	"M. P. Bends{\o}e",
	title	=	"Optimal shape design as a material distribution problem",
	journal =   "Structural Optimization",
	volume  =	1,
	number	=	4,
	pages   =   "193--202",
	month   =  "",
	year	=	1989}

@article{1992_SIMP,
	author	=	"G. I. N. Rozvany and M. Zhou and T. Birker",
	title	=	"Generalized shape optimization without homogenization",
	journal =   "Structural Optimization",
	volume  =	4,
	number	=	"3--4",
	pages   =   "250--252",
	month   =  "",
	year	=	1992}

@article{2016_lazarov,
  title={Length scale and manufacturability in density-based topology optimization},
  author={Lazarov, Boyan S and Wang, Fengwen and Sigmund, Ole},
  journal={Archive of Applied Mechanics},
  volume={86},
  number={1},
  pages={189--218},
  year={2016},
  publisher={Springer}}

@article{2019_jewett,
  title={Topology-optimized design, construction and experimental evaluation of concrete beams},
  author={Jewett, Jackson L and Carstensen, Josephine V},
  journal={Automation in Construction},
  volume={102},
  pages={59--67},
  year={2019},
  publisher={Elsevier}
}

@article{smith2024reducing,
  title={Reducing embodied carbon with material optimization in structural engineering practice: Perceived barriers and opportunities},
  author={Smith, Margaret SI and Fang, Demi and Mueller, Caitlin and Carstensen, Josephine V},
  journal={Journal of Building Engineering},
  volume={95},
  pages={109943},
  year={2024},
  publisher={Elsevier}
}

@article{loos2022towards,
  title={Towards intentional aesthetics within topology optimization by applying the principle of unity-in-variety},
  author={Loos, Shannon and Wolk, Sytze van der and Graaf, Nina de and Hekkert, Paul and Wu, Jun},
  journal={Structural and Multidisciplinary Optimization},
  volume={65},
  number={7},
  pages={185},
  year={2022},
  publisher={Springer}
}

@article{saadi2023generative,
  title={Generative design: reframing the role of the designer in early-stage design process},
  author={Saadi, Jana I and Yang, Maria C},
  journal={Journal of Mechanical Design},
  volume={145},
  number={4},
  pages={041411},
  year={2023},
  publisher={American Society of Mechanical Engineers}
}

@article{eschenauer2001topology,
  title={Topology optimization of continuum structures: a review},
  author={Eschenauer, Hans A and Olhoff, Niels},
  journal={Appl. Mech. Rev.},
  volume={54},
  number={4},
  pages={331--390},
  year={2001}
}

@article{roy2008recent,
  title={Recent advances in engineering design optimisation: Challenges and future trends},
  author={Roy, Rajkumar and Hinduja, Srichand and Teti, Roberto},
  journal={CIRP annals},
  volume={57},
  number={2},
  pages={697--715},
  year={2008},
  publisher={Elsevier}
}

@article{sigmund2013topology,
  title={Topology optimization approaches: A comparative review},
  author={Sigmund, Ole and Maute, Kurt},
  journal={Structural and multidisciplinary optimization},
  volume={48},
  number={6},
  pages={1031--1055},
  year={2013},
  publisher={Springer}
}










\end{document}